\documentclass[lettersize,journal]{IEEEtran}

\usepackage{graphicx}
\usepackage{cite}
\usepackage{picinpar}
\usepackage{amsmath}
\usepackage{url}
\usepackage[utf8]{inputenc}
\usepackage{colortbl}
\usepackage{soul}
\usepackage{multirow}
\usepackage{pifont}
\usepackage{color}
\usepackage[dvipsnames]{xcolor}
\usepackage{alltt}
\usepackage{hyperref}
\usepackage{enumerate}
\usepackage{siunitx}
\usepackage{breakurl}
\usepackage{epstopdf}
\usepackage{pbox}

\usepackage[ruled,linesnumbered, lined, noend]{algorithm2e}
\usepackage{booktabs}
 
\usepackage{bm}
\usepackage{hyperref}
\usepackage{amssymb}
\usepackage{romannum}
\usepackage{bbm}
\usepackage[caption=false,font=footnotesize]{subfig}

\usepackage{booktabs} 
\usepackage{adjustbox}
\usepackage{siunitx} 

\usepackage[normalem]{ulem}

\hypersetup{hidelinks} 
\usepackage[switch]{lineno}

\newcommand{\blue}[1]{{\color{blue} #1}}

\usepackage{siunitx}

\usepackage[multiple]{footmisc}

\begin{document}
\pagenumbering{arabic}

\title{
Bimanual Deformable  
Bag 
Manipulation
Using a Structure-of-Interest Based Neural Dynamics Model
}  

\author{
Peng Zhou$^{1}$,
Pai Zheng$^{2}$, \IEEEmembership{Senior Member,~IEEE},
Jiaming Qi$^{1}$,
Chengxi Li$^{2}$, 
Samantha Li$^{2}$,
Yipeng Pan$^{1}$,  
Chenguang Yang$^{3}$, \IEEEmembership{Fellow,~IEEE},
David Navarro-Alarcon$^{2}$, \IEEEmembership{Senior Member,~IEEE},
and Jia Pan$^{1}$, \IEEEmembership{Senior Member,~IEEE}
\thanks{
This work is supported by the Innovation and Technology Commission of the HKSAR Government under the InnoHK initiative.
\emph{(Corresponding author: Jia Pan.)}
}
\thanks{$^{1}$The University of Hong Kong, HK, Hong Kong. 
}%
\thanks{$^{2}$The Hong Kong Polytechnic University, KLN, Hong Kong. 
}%
\thanks{$^{3}$The University of Liverpool, England, UK. 
}%

}

\markboth{IEEE/ASME TRANSACTIONS ON MECHATRONICS, PREPRINT VERSION. ACCEPTED OCTOBER, 2024}
{P. Zhou \MakeLowercase{\textit{et al.}} }

\maketitle

\begin{abstract}
The manipulation of deformable objects by robotic systems presents significant challenges due to their complex dynamics and infinite-dimensional configuration spaces. This paper introduces a novel approach to deformable object manipulation (DOM) by emphasizing the introduction and manipulation of structures of interest (SOIs) in deformable fabric bags. We propose a bimanual manipulation framework that leverages a graph neural network (GNN)-based neural dynamics model to succinctly represent and predict the behavior of these SOIs. Our approach involves global particle sampling process to
construct a particle representation from partial point clouds of the SOIs and learning the neural dynamics model that effectively captures the essential deformations of the SOIs for fabric bags. By integrating this neural dynamics model with model predictive control (MPC), we enable robotic manipulators to perform precise and stable manipulation tasks focused on the SOIs. We validate our new framework through various experiments that demonstrate its efficacy in manipulating deformable bags and T-shirts. Our contributions not only address the complexities inherent in DOM but also provide new perspectives and methodologies for enhancing robotic interactions with deformable materials by concentrating on their critical structural elements.
Videos of the conducted experiments can be seen at \blue{\url{https://sites.google.com/view/bagbot}}.
\end{abstract}


\IEEEpeerreviewmaketitle


\section{Introduction}
\IEEEPARstart{D}{eformable} object manipulation (DOM) \cite{yin2021modeling, zhu2022challenges, hu20193} is a fundamental capability for robots to meaningfully interact with the physical world and assist in various human tasks. However, the manipulation of deformable objects such as cloths \cite{garcia2022household}, ropes \cite{huo2022keypoint}, and food materials \cite{lin2022diffskill} is particularly challenging due to their infinite-dimensional configuration space and complex dynamics. Traditional methods in DOM often resort to simplified physics models or data-driven modeling with handcrafted features, which lack adaptability for the varied shapes and dynamics of these objects. 
Moreover, most current DOM works focus on manipulating the entire object, neglecting the critical structures, \textit{i.e.}, \textit{structures of interest} (SOIs), that are essential for subsequent manipulation steps.

\begin{figure}[htbp]
  \begin{center}
    \includegraphics[width=\columnwidth]{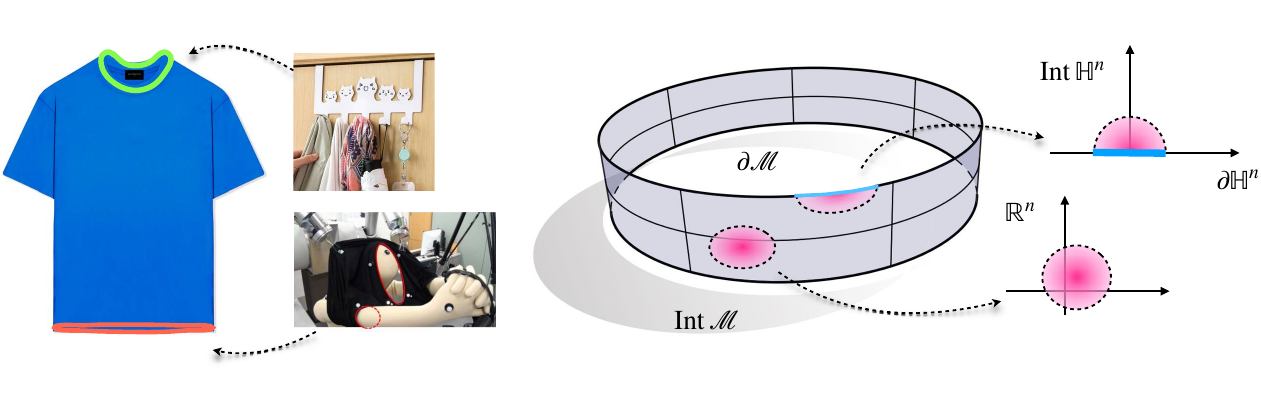
}
  \end{center} 
  \vspace{-0.6cm}
  \caption{
  (Left) SOI Examples for different deformable object manipulation tasks, e.g., garment hanging, and robot-assistive dressing.
  (Right) Conceptual representation of the manifold with boundary. The manifold encompasses $\operatorname{Int} \mathcal{M}$ and $\partial \mathcal{M}$, where the local neighborhoods of points in $\operatorname{Int} \mathcal{M}$ and ${\partial \mathcal{M}}$ are homeomorphically equivalent to $\operatorname{Int} \mathbb{H}^{n}$ and $\partial \mathbb{H}^{n}$.
  }
  \vspace{-0.4cm}
  \label{fig_soi_manifold}
\end{figure}

\begin{figure}[htbp]
    \includegraphics[width=\columnwidth]{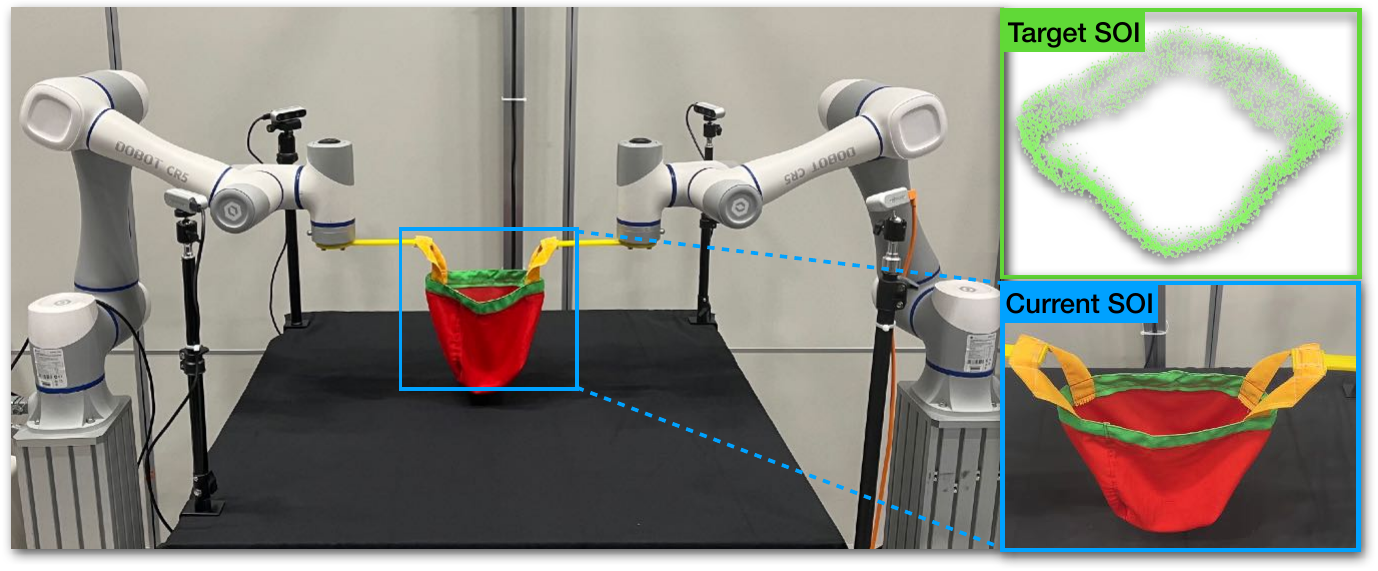}
      \vspace{-0.6cm}
  \caption{
  The two robots grasp two handles of a fabric bag to manipulate the SOI (i.e., the opening rim) into the target configuration. 
  }
  \vspace{-0.5cm}
  \label{fig_example}
\end{figure}

In this paper, we introduce the concept of SOI into the realm of DOM (see Fig. \ref{fig_soi_manifold} for an example), a paradigm shift that emphasizes the importance of leveraging and manipulating key structural components rather than the entire object. This focus on the SOI is motivated by the observation that successful DOM tasks typically involve the manipulation of these key areas. By targeting these SOIs, we can reduce the computational load significantly, as modeling the complete 3D dynamics of the deformable object is unnecessary and burdensome for the task at hand.

In this work, we address the bimanual manipulation of a deformable fabric bag as illustrated in Fig. \ref{fig_example}. Our proposed framework introduces a GNN-based neural dynamics model that enables to tackle the challenges associated with deformable object manipulation (DOM), with an emphasis on Structures of Interest (SOIs). The framework seeks to effectively represent the state of the object by extracting SOIs' particles from the point cloud data, and learning the underlying dynamics within a condensed particle space. By integrating this method with model predictive control (MPC), we enable robots to achieve accurate and stable manipulation of deformable bags. The contributions of this work are as follows:
\begin{itemize}
	\item The development of an original bimanual manipulation strategy for deformable fabric bags with an emphasis on the utility of SOIs.
	\item The adoption of the SOI concept for deformable object representations using the proposed global particle sampling method, highlighting the significance of key structural elements during manipulation.
	\item A new approach to learning the SOI-based neural dynamics models via GNNs, applied to particle sets of SOIs.
	\item The application of MPC informed by the neural dynamics model to guide the generation of optimal manipulation actions that prioritize SOIs.
\end{itemize}

Our experiments demonstrate the potential of this framework in the context of bimanual  manipulating deformable objects, such as fabric bags, T-shirts. While our findings suggest improvements in the robotic manipulation of deformable objects by concentrating on SOIs, we acknowledge the ongoing need for research in this area. This work contributes to the broader understanding of intelligent manipulation strategies and opens pathways for future innovations in the robotic handling of complex materials.

 
\section{Related Work}
Deformable object manipulation (DOM) has been an active research area in robotics. Existing methods can be categorized into model-based and data-driven approaches.
Model-based methods rely on simplified physics models to represent deformable objects. Early works used mass-spring models (MSM) to simulate deformation \cite{makiyeh2022indirect}. The finite element method (FEM) provides more accurate modeling of continuum mechanics \cite{zhang2017visual, ficuciello2018fem}. However, analytical models require extensive manual tuning and generalization across different materials or shapes remains difficult.
Data-driven methods \cite{xu2022dextairity, zhou2024reactive} aim to learn models directly from data. Vision-based methods extract geometric features from visual observations to infer deformations \cite{navarro2016automatic, qi2021contour}. Recent works utilized deep learning on point cloud data and achieved improved modeling accuracy \cite{lin2022planning}. However, they depend heavily on large labelled datasets. Self-supervised methods were proposed to learn from physical interactions \cite{nair2017combining, yan2020self}. However, they focused on planar objects and could not handle complex deformations.

Recent advancements in DOM have leveraged graph neural networks (GNNs) to model the complex interactions within varying materials and shapes \cite{gasteiger2021gemnet, zhou2024imitating, tolstaya2020learning, bertiche2022neural}. Wang \textit{et al.} \cite{wang2022offline} proposed an offline-online learning framework that utilizes GNNs for the deformation model in cable manipulation. This method captures the intricate physics of cables, achieving a balance between offline learning from a rich dataset and online refinement through interaction. Deng \textit{et al.} \cite{deng2022deep} introduced a deep reinforcement learning approach that employs local GNNs for a goal-conditioned rearranging of deformable objects. Their method efficiently adapts to the changing dynamics of the objects during manipulation tasks. In the domain of elasto-plastic object manipulation, Shi \textit{et al.} \cite{shi2023robocraft} presented RoboCraft, which combines perception and simulation through GNNs to understand and shape 3D objects. Furthermore, the GDOOM framework by Ma \textit{et al.} \cite{ma2022learning} focuses on learning latent graph dynamics for visual manipulation, offering a comprehensive approach to predicting deformable object behavior.

In the specific area of deformable bag manipulation, a number of approaches have been proposed. Xu \textit{et al.} \cite{xu2022dextairity} developed DextAIRity, a system that simplifies the manipulation of deformable objects by harnessing pneumatic controls. Gu \textit{et al.} \cite{gu2024shakingbot} introduced ShakingBot, which demonstrates dynamic manipulation skills for bagging tasks, highlighting the potential to handle high-speed and complex deformable object interactions. Chen \textit{et al.} \cite{chen2023autobag} focused on the challenge of opening plastic bags and inserting objects, a task with significant practical applications in automation. Lastly, Weng \textit{et al.} \cite{weng2024interactive} explored the concept of interactive perception for deformable object manipulation, where the perception and interaction are tightly integrated to enhance manipulation capabilities.

Our work distinguishes itself from the existing literature by introducing a novel bimanual manipulation framework for deformable fabric bags that places a strong emphasis on \textit{structures of interest} (SOIs). Unlike previous work that focused on the manipulation of entire objects or specific applications such as fabric or plastic bag handling, our approach centers on identifying and manipulating the SOIs within fabric bags to achieve precise and efficient DOM. We propose a GNN-based latent dynamics model tailored to represent and predict the behavior of these SOIs, which is a subtle yet meaningful advancement from the general-purpose models in the current literature. Furthermore, by integrating this model with model predictive control (MPC), our framework enables accurate and stable manipulation of fabric bags by robotic systems in a computationally efficient manner. Our contributions lie in the specialized focus on SOIs and the combination of GNN-based modeling with MPC, subtly enriching the landscape with fresh viewpoints for enhancing robotic DOMs.

\begin{figure}[t]
	\centering
	\includegraphics[width=\columnwidth
	]{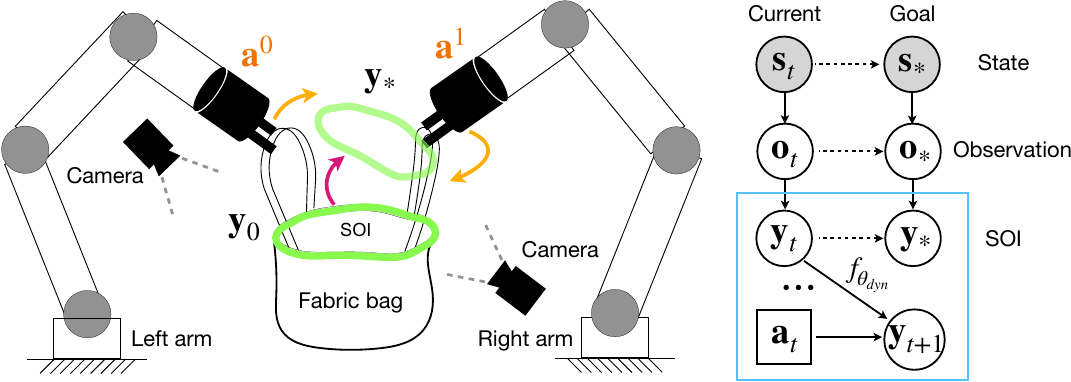}
	\vspace{-0.8cm}
	\caption{
(Left) Conceptual representation of SOI-based bimanual deformable object manipulation problem; (Right)	The problem is formulated as a POMDP setting, where the SOI-related observation $\mathbf{o}^{soi}_t$ is extracted from the original observation $o_t$ and governed by $f_{\theta_\mathrm{dyn}}$.
	}
	\label{fig_pomdp}
	\vspace{-0.2cm}
\end{figure}

\section{Problem Statement}
Given that individual image and depth observations generally do not fully disclose the state of the environment, we approach the task of bimanual bag manipulation as a Partially Observable Markov Decision Process (POMDP) as depicted in Fig. \ref{fig_pomdp}. This is formally defined by the tuple \((\mathcal{S}, \mathcal{A}, \mathcal{T}, \mathcal{O}, \Omega, \mathcal{R}, \gamma)\), where the state at time \(t\), denoted by \(s_t\), belongs to the state space \(\mathcal{S}\) and is not directly observable. The state encapsulates the configuration of the robots and the manipulated deformable object. The corresponding observation at time \(t\), denoted by \(o_t\), is within the observation space \(\mathcal{O}\).
The state transition model \(\mathcal{T}(s_{t+1} \mid s_t, a_t)\) describes the probability of transitioning from the current state \(s_t\) to a new state \(s_{t+1}\) upon taking an action \(a_t\) from the action space \(\mathcal{A}\), which consists of the combined left and right robotic actions, represented by the Cartesian product \(\mathcal{A} = \mathcal{A}^{0} \times \mathcal{A}^{1}\).
The function \(\Omega(o_t \mid s_t, a_{t-1})\) specifies the likelihood of observing \(o_t\) after executing action \(a_{t-1}\) and transitioning to state \(s_t\).
The reward function \(\mathcal{R}(s_t, a_t)\) assigns a valued reward to each state-action pair, and the discount factor \(\gamma \in [0,1)\) quantifies the preference for immediate rewards over future rewards.

The goal of this work is to use two robot manipulators to firmly grasp and manipulate the SOI of the deformable object to achieve a target SOI state $\mathbf{y}_*$.
We assume this deformable object manipulation task is a quasi-static manipulation and dynamic manipulation motions are not considered.
As shown in Fig. \ref{fig_pomdp}, at time step $t$, the dual-arm manipulators apply action $\{ \mathbf{a}^0_t, \mathbf{a}^1_t \} = \mathbf{a}_t \in \mathcal{A}$ upon the deformable object, and we can partially observe transitions of the object from $\mathbf{o}_t$ to $\mathbf{o}_{t+1}$ under the unknown state transitions from $\mathbf{s}_t$ to $\mathbf{s}_{t+1}$.
However, a complete observation of the bag is not necessary, in our task, the opening rim of the bag is critical for successful manipulation tasks since it not only determines the manipulation task goals but also provides the most informative sensory feedback, such as visual landmarks, during manipulation.
We define the structure of Interest (SOI) in the context of deformable object manipulation (DOM) refers to specific regions or features of a deformable object that are critical for successful manipulation tasks (\textit{e.g.}, examples in Fig. \ref{fig_soi_manifold}). 
Therefore, in this task, we consider the opening rim of the manipulated bag as our SOI points, and topologically, we can define this loop-like structure as a \textit{manifold with boundary} \cite{lee2012smooth}. As illustrated in Fig. \ref{fig_soi_manifold}, we also define its $\textit{interior}$ and $\textit{boundary}$ as $\operatorname{Int} \mathcal{M}$ and ${\partial \mathcal{M}}$, whose points' neighborhoods are respectively \textit{homeomorphic} to $\operatorname{Int} \mathbb{H}^{n} =\{(x_{1}, \ldots, x_{n}) \mid x_{n}>0\}$ and
$\partial \mathbb{H}^{n} =\{(x_{1}, \ldots, x_{n})\mid x_{n}=0\}$.

With an appropriate perception module, the observation of the SOI, denoted by \(\mathbf{o}^{soi}_t\), can be extracted from the overall observation of the bag \(\mathbf{o}_t\). Our approach is based on the insight that it is more efficient to predict the dynamics of the SOI rather than the entire complex dynamics of the bag. To this end, we employ a graph neural network (GNN) to establish a dynamics model \(f_{\theta_{\text{dyn}}}\) that is dedicated to learning the transition functions of the SOI, defined as \(f_{\theta_{\text{dyn}}}: \mathcal{O}^{soi} \times \mathcal{A} \rightarrow \mathcal{O}^{soi} \).
This dynamics model accepts as input a sequence of SOI observations $\mathbf{o}^{soi}_{t-n:t} \in \mathcal{O}^{soi}$ and bimanual actions $\{\mathbf{a}^0, \mathbf{a}^1\}_{t-n: t} \subseteq \mathcal{A}$, and predicts the subsequent SOI observation $\mathbf{o}^{soi}_{t+1}$, where \(n\) represents the length of the observation history before the current time step \(t\). 
With the dynamics model, we proceed to cast the bimanual manipulation of the bag as a task within the model predictive control (MPC) framework.
Within this MPC setup, the cost function \(\mathcal{J}\) quantifies the difference between the final SOI observation at time step \(T\) and the targeted SOI state $\mathbf{o}^{soi}_*$.
Details on the precise structure of the cost function \(\mathcal{J}\) are illustrated
in the following section. This cost is minimized to yield an optimal sequence of actions across a temporal horizon of \(T\) steps:
\begin{equation}
\begin{aligned}
		\{ \mathbf{a}^0, \mathbf{a}^1 \}_{0:{T-1}} &= \underset{ \{ \mathbf{a}^0, \mathbf{a}^1 \}_{0:T-1}  \subseteq \mathcal{A}}{\arg \min} \mathcal{J}(\mathbf{o}^{soi}_T, \mathbf{o}^{soi}_*)\\
	\text{where}	 \quad \mathbf{o}^{soi}_T &= f_{\theta_{dyn}} \left(\mathbf{o}^{soi}_0, \{ \mathbf{a}^0, \mathbf{a}^1 \}_{0:{T-1}}\right)
\end{aligned}
\end{equation}
where \(\mathbf{o}^{soi}_T\) represents the predicted SOI state at time step \(T\).

\begin{figure*}[htbp]
    \includegraphics[width=2\columnwidth]{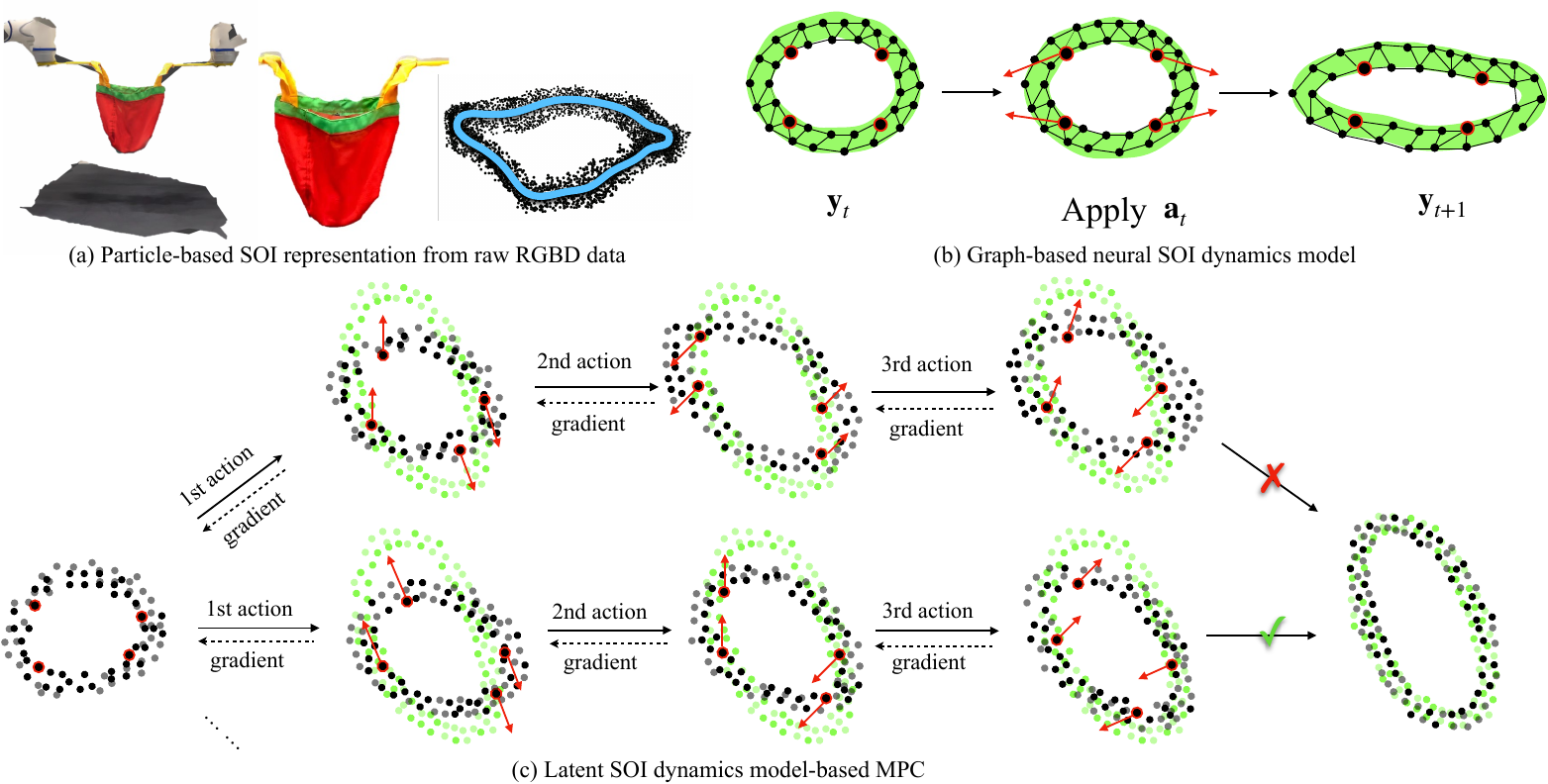}
  \caption{
  The conceptual representation of the proposed framework for bimanual deformable fabric bag manipulation using the latent SOI dynamics model.
  }
  \label{fig_framework}
  \vspace{-0.4cm}
\end{figure*}

\begin{figure}[htbp]
  \includegraphics[width=1\columnwidth]{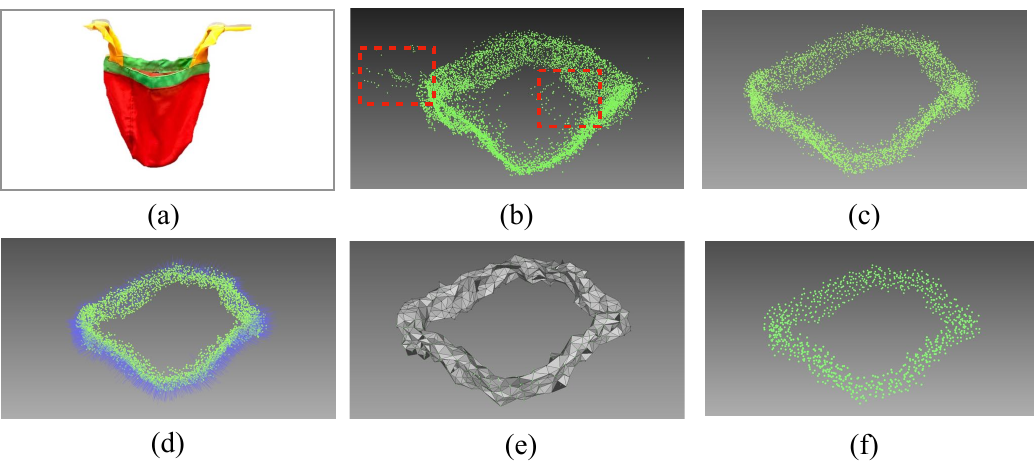}
  \vspace{-0.6cm}
  \caption{
The proposed global particle sampling process:
(a) Reconstructed point cloud (PCD) $\mathcal{P}{\text{full}}$;
(b) SOI point cloud after HSV filtering $\mathcal{P}{\text{soi}}$;
(c) SOI point cloud after down-sampling and outlier removal;
(d) Normal generation;
(e) Reconstructed SOI surface using ball pivoting;
(f) SOI particles after uniform sampling.
}
  \label{fig_process}
  \vspace{-0.6cm}
\end{figure}

\section{Methodology}
In this section, we detail our proposed framework for bimanual deformable bag manipulation using a structure-of-interest (SOI) based neural dynamics model. We begin by extracting SOI particles from RGBD data using global particle sampling approach to represent critical bag features, as depicted in Fig. \ref{fig_framework}(a). A Graph Neural Network (GNN) then models these particles' dynamics for state prediction (Fig. \ref{fig_framework}(b)), which informs the Model Predictive Control (MPC) to adjust robotic gripper actions for shaping the SOI (Fig. \ref{fig_framework}(c)). This integrated approach streamlines the manipulation process, ensuring target SOI configurations are achieved with precision.

\begin{algorithm}[htbp]
\label{alg_hog}
\caption{Global Particle Sampling} 
\LinesNumbered
\KwIn{
SOI observation at time step $t$, $\mathbf{o}^{soi}_t$, composed of
point cloud data $\mathcal{P}^1$, $\mathcal{P}^2$, $\mathcal{P}^3$ and $\mathcal{P}^4$ from four different perspectives.
}
\KwOut{
SOI particle set $\mathbf{P}_t$ at time step $t$
}
\For {\textit{each time step $t$}}
{
	$\mathcal{P}_{full}$ $\leftarrow$ point-cloud registration based on $\mathbf{o}^{soi}_t$ \;
	$\mathcal{P}_{soi}$ $\leftarrow$ extract SOI using $\operatorname{HSV\_filtering}(\mathcal{P}_{full})$\;
	$\mathcal{P}_{soi}$ $\leftarrow$ downsampling using $\operatorname{VGD}(\mathcal{P}_{soi}, v_s)$ \;
	$\mathcal{P}_{soi}$ $\leftarrow$ remove outliers using $\operatorname{SOR}(\mathcal{P}_{soi}, k)$ \;
	$\mathcal{M}_{soi} \leftarrow$ reconstruct surface using $\operatorname{ball\_pivoting}$\;
	$\mathbf{P}_t$ $\leftarrow$ sample particles using $\operatorname{uniform\_sampling}$ $(\mathcal{M}_{soi})$ \;	
	\Return SOI particle set $\mathbf{P}_t$ \;
}
\end{algorithm}
\vspace{-0.2cm}

%
%

\subsection{Global Particle Sampling from Raw Observation}


To address the challenge of real-time extracting meaningful particle representations of the SOI from significantly occluded visual data, we introduce a comprehensive global particle sampling approach, which is illustrated in Fig. \ref{fig_process}.

\textbf{Preprocessing}:
The initial phase of our methodology involves preprocessing of the raw point cloud data obtained from RealSense D435 cameras. 
With appropriate calibration, we can transform the raw RGB-D images into point cloud data (PCD) (see Fig. \ref{fig_process}(a)) with the cameras' intrinsic and extrinsic parameters.
Then, we focus on the SOI (i.e., opening rim) of the fabric bag and employ a color-based segmentation algorithm by performing a specific color filtering within the HSV color space to isolate the rim from the rest of the bag (see Fig. \ref{fig_process}(b)). 
To enhance the processing efficiency in the following steps, we further reduce the points to satisfy the real-time performance using a parallel Voxel Grid Downsampling (VGD) by dividing the point cloud into a regular grid of 3D voxels (voxel size denoted by $v_s$) and replaces all the points within each voxel with their centroid or average.
Then, followed by a GPU-accelerated implementation of the Statistical Outlier Removal (SOR) with a $k$-nearest neighbors to remove the noise and further improve the data quality (see Fig. \ref{fig_process}(c)).
In general, this step is intended to significantly reduce the size and complexity of the point cloud to an appropriate scale (typically, around 500 points) for subsequent surface reconstruction.


\textbf{Surface Reconstruction and Refinement}:
To reconstruct the surface of SOI from its point cloud in a high efficiency, we leverage ball pivoting algorithm that is a preferred technique for its efficiency in managing point clouds.
To achieve real-time surface reconstruction, we parallelize the algorithm, which allows the BPA to utilize multiple cores or processors simultaneously, thereby accelerating the reconstruction process.
Specifically, we employ an Octree-based data structure to partition the point cloud into smaller subsets, each of which can be processed independently.
For example, we can compute point normal (see Fig. \ref{fig_process}(d)) via Octree-based nearest-neighbor search methods in each partitioned subset. 
Then we refine the reconstructed surface based on topological priors to get rid of \textit{non-manifold} edges and vertices which can be easily detected by Open3D\footnote{https://www.open3d.org/docs/release/tutorial/geometry/mesh.html}.
\textit{Non-manifold} edges and vertices in a mesh are those that do not comply with the manifold criteria, which typically require that each edge belong to exactly two faces and each vertex to form a closed fan of faces around it.
 By removing non-manifold edges and vertices, we can ensure a well-defined volume to further improve the following resampling accuracy.
 

\textbf{Resampling}:
In the final stage of our process, our objective is to generate a high-quality particle set that accurately represents the geometry of the SOI-based on the reconstructed mesh. 
To achieve this, we employ uniform sampling techniques to selectively reduce the SOI point cloud in to an appropriate fixed scale, maintaining a uniform distribution of the generated particles. 
This ensures a manageable quantity conducive to Graph Neural Network (GNN) training. The computation of this process, resulting in a refined particle dataset, is depicted in Fig. \ref{fig_process}(f).

\subsection{SOI-based Neural Dynamics Model}\label{gnn_dyn}
To characterize the dynamics of SOI particles, as illustrated in Fig. \ref{fig_framework}(b), we begin to construct a particle graph and introduce the graph neural networks to model the SOI dynamics. To construct a particle graph, 
 we denote a graph based the SOI observation to represent the SOI state as $ \mathbf{y}_t = \mathcal{G}(\mathbf{o}^{soi}_t) = (\mathbf{P}_t, \mathbf{E}_t)$, and the graph vertex set $\mathbf{P}_t$ correspond to the SOI's particles $\mathbf{p}_{i, t}$. 
 Each particle is expressed as $\mathbf{p}_{i, t}=\langle\mathbf{x}_{i, t}, \mathbf{b}_{i, t}\rangle$, where the position and attributes of the particle $i$ at time $t$ are denoted as $\mathbf{x}_{i, t}$ and $\mathbf{b}_{i, t}$, respectively. 
Attributes of the particles $\mathbf{b}_{i, t}$ are classified into handle particles and non-handle particles. 
The handle particles are identified by a open ball function as below:
\begin{equation}
\mathcal{B}_r(\mathbf{x}_{handle}) =\{ \mathbf{p}_i \in \mathbf{P} ~|~ d(\mathbf{p}_i, \mathbf{x}_{handle})<r\}	
\end{equation} 
where $d(\mathbf{p}_i, \mathbf{x}_{handle})$ represents the distance between SOI particles and bag handle position in Euclidean space.
The handle position is also identified by HSV filtering in color space.
Introducing this particle attribute significantly enhances the SOI neural dynamics model. The neural dynamics model relies on understanding how different parts (nodes) of the SOI, which lie on the deformable object, behave under robot manipulations. By identifying the position of the handle, which has a strong relationship with the enforced robot actions, the model can accurately locate and classify these critical nodes.
Therefore, node classification allows the model to learn and predict the dynamics of specific particles of the SOI, particularly those near the handle, with greater accuracy.
 Edges $\mathbf{E}_t$ link vertices dynamically, based on spatial relationships, connecting all neighboring particles within a predefined range. Edge relations are captured by $\mathbf{e}_j=\langle m_j, n_j, \mathbf{c}_j\rangle$, with $1 \leq m_j, n_j \leq |\mathbf{P}_t|$ denoting the indices of the connected particles, $j$ being the edge index, and $\mathbf{c}_j$ describing the type of connection, be it internal structural or handle-to-rim linkages.

The goal of constructing a forward dynamics model is to predict SOI subsequent state $\mathbf{y}_{t+1}$ based on a short historical sequence of SOI states, denoted by:
\begin{equation}
		\mathbf{y}_{t+1} = f_{\theta_{\text{dyn}}} \left( \mathbf{y}_{t-h:t}, \mathbf{a}_t \right) 
\end{equation}
With the application of particle graph to represent the SOI states, 
we can introduce GNNs to simulate the neural dynamics of the SOI to predict subsequent states from a short historical sequence of SOI 
 particle graphs, extended as:
\begin{equation}
\mathcal{G}(\mathbf{o}^{soi}_{t+1}) = f_{\theta_{\text{dyn}}} \left(\mathcal{G}(\mathbf{o}^{soi}_{t-h:t}), \{ \mathbf{a}^0, ~\mathbf{a}^1 \}_t \right)
\end{equation}
To facilitate this, the original high-dimensional observation space concerning the SOI is reduced into a condensed, low-dimensional latent graph by encoding the distinct particle and connection features of the SOI. 
The loss function of each encoder is defined based on a distance metric as:
\begin{equation}
\begin{aligned}
\phi^\mathbf{p}, \psi^\mathbf{p} &= \arg\min_{\phi^\mathbf{p}, \psi^\mathbf{p}}\operatorname{Dist}[\mathbf{p}_i - (\phi^\mathbf{p} \circ \psi^\mathbf{p})(\mathbf{p}_i) ] \\
\phi^\mathbf{e}, \psi^\mathbf{e} &= \arg\min_{\phi^\mathbf{e}, \psi^\mathbf{e}}\operatorname{Dist}[\mathbf{e}_j - (\phi^\mathbf{e} \circ \psi^\mathbf{e})(\mathbf{e}_j) ]
\end{aligned}
\end{equation}
Here, $\phi^\mathbf{p}: V \rightarrow Z_{V}$ and $\phi^\mathbf{e}: E \rightarrow Z_{E}$ serve as the encoders for SOI particles and edges, while $\psi^\mathbf{p}: Z_{V} \rightarrow V$ and $\psi^\mathbf{e}: Z_{E} \rightarrow E$ function as their respective decoders. The encoding of SOI particle and edge features via $\phi^\mathbf{p}$ and $\phi^\mathbf{e}$ is executed as follows:
\begin{equation}
\begin{aligned}
z_{i, t}^{\mathbf{p}} &= \phi^\mathbf{p}(\mathbf{p}_{i, t}) \\
z_{j, t}^{\mathbf{e}} &= \phi^\mathbf{e}(\mathbf{p}_{m_j, t}, \mathbf{p}_{n_j, t}, \mathbf{c}_j)
\end{aligned}
\end{equation}
Subsequently, the dynamics are captured using the decoders $\psi^\mathbf{p}$ and $\psi^\mathbf{e}$, leading to the prediction of the SOI particle graph at time $t+1$:
\begin{equation}
\begin{aligned}
\mathbf{\hat{e}}_{j, t} &= \psi^\mathbf{e}(z_{j, t}^{\mathbf{e}})_{j=1, \cdots, |\mathbf{E}_t|} \\
\hat{\mathbf{p}}_{i, t+1} &= \psi^\mathbf{p}\left(z_{i, t}^{\mathbf{p}}, \sum_{j \in \mathcal{N}_i} \mathbf{\hat{e}}_{j, t} \right)_{i=1, \cdots, |\mathbf{P}_t|}
\end{aligned}
\end{equation}
Within this context, $\mathcal{N}_i$ denotes the set of edges where particle $i$ is the recipient. To adeptly handle the instantaneous propagation of forces, the training also integrates multistep message passing.

Since our training data is based on the particle sets generated from raw point cloud data with our proposed global particle sampling process, it lacks a consistent point-to-point mapping across each frame. Therefore, to quantify the similarity between two sets of SOI particle distributions, we investigate two permutation-invariant loss functions. 
The first is the widely adopted Chamfer distance (CD), computed between two particle sets $\mathbf{P}_1$ and $\mathbf{P}_2$ as follows:
\begin{equation}
\small
\mathcal{L}_{\text{CD}}(\mathbf{P}_1, \mathbf{P}_2)=\sum_{\mathbf{x}_1 \in \mathbf{P}_1} \min_{\mathbf{x}_2 \in \mathbf{P}_2}\|\mathbf{x}_1-\mathbf{x}_2\|_2^2 + \sum_{\mathbf{x}_2 \in \mathbf{P}_2} \min_{\mathbf{x}_1 \in \mathbf{P}_1}\|\mathbf{x}_1-\mathbf{x}_2\|_2^2
\end{equation}
The second is the Earth mover's distance (EMD), which is formulated as:
\begin{equation}
\small
\mathcal{L}_{\text{EMD}}(\mathbf{P}_1, \mathbf{P}_2)=\min_{\lambda : \mathbf{P}_1 \rightarrow \mathbf{P}_2} \sum_{\mathbf{x}_1 \in \mathbf{P}_1}\|\mathbf{x}_1-\lambda(\mathbf{x}_1)\|_2
\end{equation}
where $\lambda : \mathbf{P}_1 \rightarrow \mathbf{P}_2$ denotes a bijective mapping. 
The EMD represents a solution to the assignment problem, ensuring a unique and stable optimal bijection $\lambda$ for almost every pair of particle sets, invariant to infinitesimally small point displacements. 
Note that the bijection $\lambda$ will be computed between two particle set for every time which means the bijection will be regenerated.
To ensure real-time performance, we will limit the cardinality of the particle set by configuring the GSP to generate a small number of particles. This reduction in particle count helps minimize the computational burden, allowing us to meet the necessary requirements for a real-time controller.
In essence, EMD in our framework aligns distributions while mitigating point cloud anomalies through the definition of bijective correspondence.
Regarding the Chamfer distance, it's worth noting that its use here is somewhat liberal, as it does not fulfil the triangle inequality property.
Our composite loss function integrates these distances in a weighted fashion: $\mathcal{L}(\mathbf{P}_1, \mathbf{P}_2)= \alpha \mathcal{L}_{\text{CD}}(\mathbf{P}_1, \mathbf{P}_2) + \beta \mathcal{L}_{\text{EMD}}(\mathbf{P}_1, \mathbf{P}_2)$. Empirical evaluations suggest that the optimal weights are $\alpha=0.15$ and $\beta=0.85$.


\subsection{Model Predictive Control}
 \begin{figure}[t!]
   \vspace{-0.3cm}
    \includegraphics[width=\columnwidth]{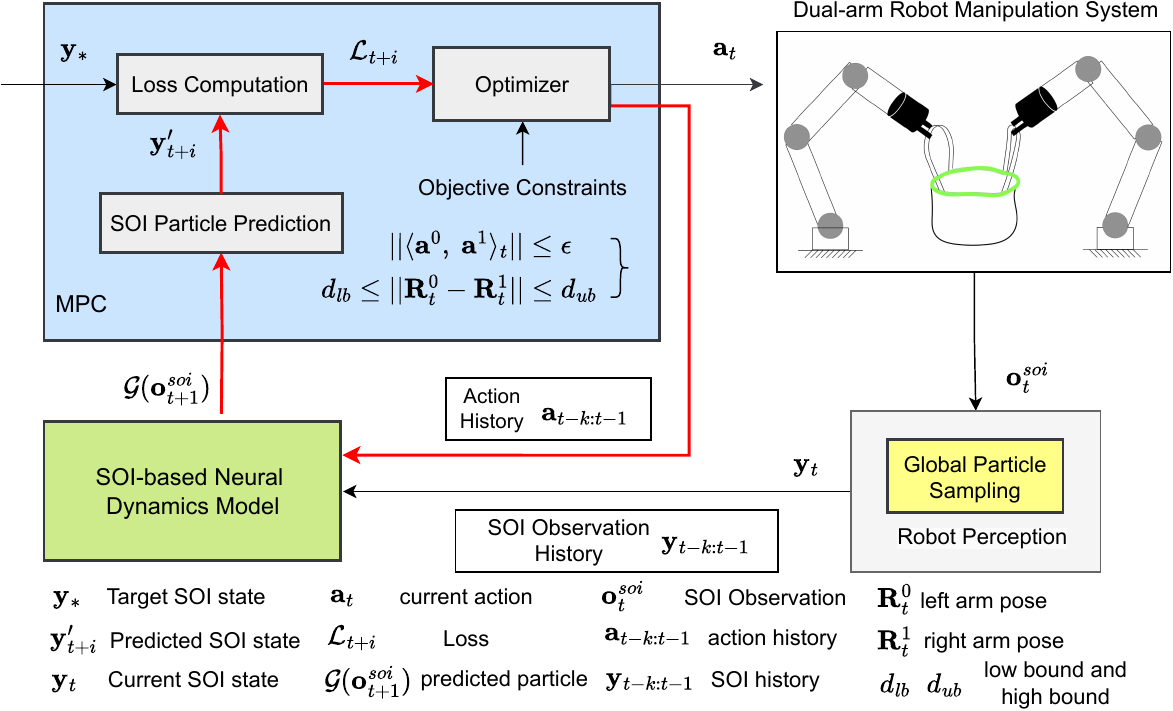}  
  \caption{
The MPC framework integrates our trained SOI-based neural dynamics model for bimanual deformable bag manipulation.
  }
  \label{fig_mpc}
\end{figure}

Upon training our SOI-centric neural dynamics model, we integrate a model predictive control (MPC) approach to control the robotic gripper in manipulating the fabric bag, as depicted in Fig. \ref{fig_mpc}. We simplify the gripper's action space into a parameterized form: $(x, y, z, r_z)$, with $\{x, y, z\}$ representing the end-effector's position interacting with the bag handles, and $r_z$ indicating the gripper's rotation around the vertical $z$ axis. 
We considered only $r_z$ in our action space based on our observation that $r_z$ induces larger SOI configuration changes compared to $r_x$ and $r_y$.
A goal-oriented MPC with $T$ as the planning horizon for the desired SOI observation $\mathbf{o}^{soi}_*$, denoted as the particle set $\mathbf{P}_{\mathbf{o}^{soi}_*}$, is employed as below:
\begin{equation}
	\begin{aligned}
	\small
\min _{\{ \mathbf{a}^0, ~\mathbf{a}^1 \}_{0:T-1}} & \sum_{t=0}^{T} \mathcal{L}\left(\mathbf{P}_{\mathbf{o}^{soi}_t}, \mathbf{P}_{\mathbf{o}^{soi}_*} \right)
\\
\text{ s.t. } & \mathcal{G}(\mathbf{o}^{soi}_{t+1}) = f_{\theta_{\text{dyn}}} \left(\mathcal{G}(\mathbf{o}^{soi}_{t-h:t}), \{ \mathbf{a}^0, ~\mathbf{a}^1 \}_t \right) \\
&  d_{lb} \leq || \mathbf{R}^0_t - \mathbf{R}^1_t || \leq d_{ub} \\
& || \langle \mathbf{a}^0, ~\mathbf{a}^1 \rangle_t || \leq \epsilon \\
\end{aligned}
\end{equation}
where we constrain the distance between left robot end-effector pose $\mathcal{R}^0$ and right robot end-effector pose $\mathcal{R}^1$ to fall within a specified range from the lower bound $d_{lb}$ to the upper bound $d_{ub}$ so as to prevent potential collisions between the manipulators and safeguard the fabric from tearing due to excessive force. Additionally, we ensure that the magnitude of the robot's actions does not exceed  $\epsilon$ to enforce a smooth and continuous trajectory.

As illustrated in Fig. \ref{fig_framework}(c), gradient-based trajectory optimization is utilized to identify the trajectory with the lowest defined cost. 
First, random shooting within the simplified actions is performed,  followed by the computation of costs using the SOI-based latent dynamics model.
Subsequently, we employ the limited-memory BFGS method \cite{liu1989limited} on the trajectories with the lowest cost to optimize the manipulation actions with gradients, using the same loss function during the dynamics model's training process.
The global convergence is guaranteed as proved in \cite{mokhtari2015global} and we empirically demonstrated the effectiveness through experiments.

\section{Experiments}
\subsection{Experiment Setup}
\begin{figure}[tb]
 \vspace{-0.3cm}
    \includegraphics[width=\columnwidth]{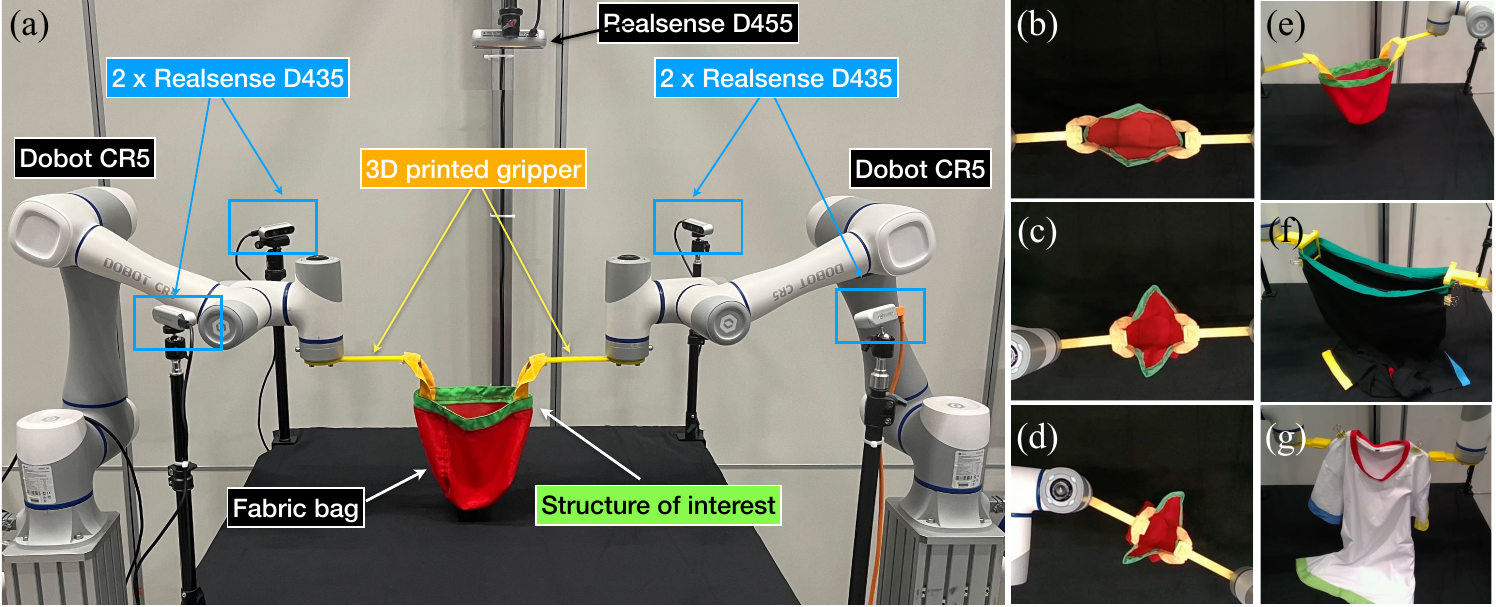}  
  \caption{
Experimental set-up to validate bimanual deformable bag manipulation tasks.
 (a) Overview of the set-up. (b)-(d) The different SOI shape categories in bag experiments: Long Oval, Round Oval, and Short Oval. (e)-(g) The considered bag SOI and two additional types of T-shirt SOIs: opening rim, bottom hem and collar.
  }
  \label{fig_exp_setup}
\end{figure}

As shown in Fig. \ref{fig_exp_setup}, we present the general experimental setup used to validate our SOI-based dynamics model for bimanual deformable bag manipulation tasks. The setup includes four RealSense D435 RGB-D cameras, each positioned at a corner, to capture RGB-D images of the deformable fabric bag from multiple angles at a frequency of 30Hz and a resolution of 640$\times$480. Two Dobot CR5 robotic manipulators, equipped with 7 degrees of freedom, grasp different SOIs using 3D-printed grippers secured with zip ties or binder clips.
To validate the effectiveness of our proposed approach, three SOIs of the deformable objects, including the opening rim of a bag, the bottom hem and the collar of a T-shirt, have been modified for better perception: it is cut and sewn with the same type of fabric but in contrasting colors, which facilitates the detection while maintaining the consistency of the SOI's dynamic behavior. Additionally, a top-down camera and a front-facing camera are employed to capture the experimental process from various perspectives.
To support real-time point cloud process, particle sampling and MPC optimization, we utilize a workstation with Ubuntu 20.04, an Intel Core i9-14900K CPU (24 cores, 24 threads), an Nvidia RTX 4090 GPU, 128GB of DDR4/DDR5 RAM, and a 2TB NVMe SSD for storage.
Table \ref{tab_compute} presents the detailed computation time for the proposed global particle sampling and GNN-based neural dynamics model-integrated MPC optimization process. The total computation time required to generate a robot action command is approximately 63.75ms, theoretically allowing the robot to execute up to 15.69 action commands per second to manipulate deformable objects.
However, to maintain consistency with the action execution frame rate used during the training of our GNN-based neural dynamics model, we standardize the execution frame rate of robot action commands to 10 FPS.

\begin{table}[tb]
\caption{
Computation time of the proposed global particle sampling and MPC optimization process.
}
\label{tab_compute}
\centering
\begin{tabular}{l S[table-format=2.2]}
 \toprule
Process & {Computation Time (ms) $\downarrow$} \\
\midrule 
Point Cloud Registration & 3.13 ~$\pm$~ 0.26 \\
SOI Extraction &   6.94 ~$\pm$~ 0.47 \\
Down-sampling & 	  7.05 ~$\pm$~ 0.73 \\         
Outlier Removal & 8.75 ~$\pm$~ 0.61 \\
Ball Pivoting &   15.65 ~$\pm$~ 1.42 \\
Particle Sampling & 6.08 ~$\pm$~ 0.37 \\
MPC Computation  & 17.12 ~$\pm$~ 1.79 \\
GNN Inference & 0.37 ~$\pm$~ 0.03 \\
\bottomrule
\end{tabular}
\end{table}

 \begin{figure*}[tbp]
\centering
    \includegraphics[width=2\columnwidth]{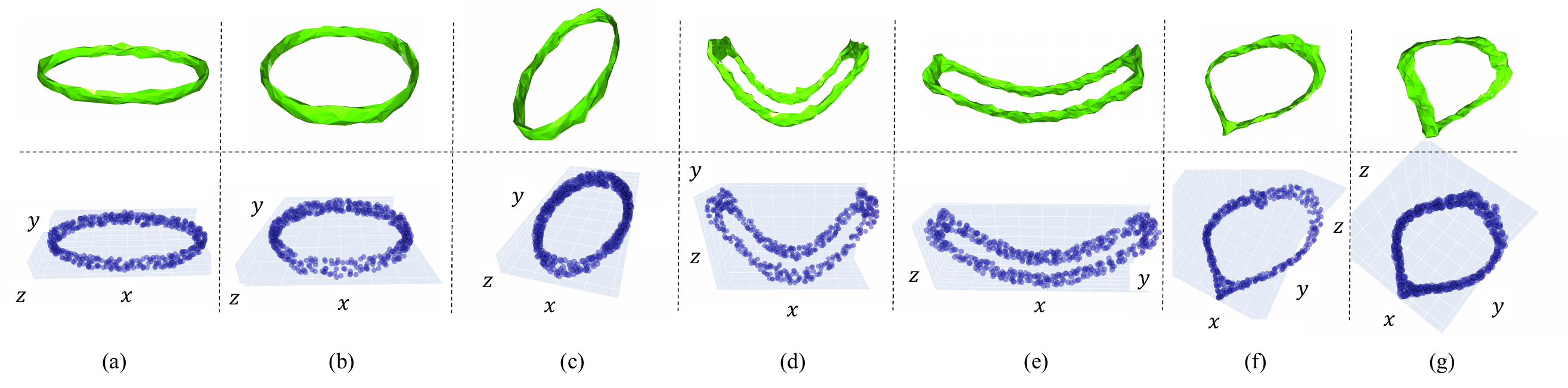}  
  \vspace{-0.3cm}
  \caption{ Results of our proposed global particle sampling approach.
(Top) Reconstructions of different SOIs for different deformable objects.
(Bottom) Corresponding SOI particle sets.
(a)-(c): bag opening rim, (d)-(e): T-shirt bottom hem, and (f)-(g): T-shirt collar.
  }
  \label{fig_particle}
\end{figure*}

As illustrated in Fig. \ref{fig_exp_setup}(b)-(d), our experiment considers three SOI shape categories for bag manipulation tasks: Long Oval (LO), Round Oval (RO), and Short Oval (SO). Additionally, we have incorporated experiments involving two additional types of T-shirt SOIs, each made from distinct fabric materials to further validate the effectiveness of our proposed approach on a variety of deformable objects. One SOI is the collar crafted from a soft modal fabric (Fig. \ref{fig_exp_setup}(g)), while the other hem SOI comprises a slightly stiffer polyester-cotton blend (Fig. \ref{fig_exp_setup}(f)). These variations in material properties provide a more comprehensive assessment of our method's adaptability and robustness across different fabric types. For each deformable object, we collect a training dataset of 72,000 frames around 120 minutes, featuring 600 episodes and each spanning 120 frames. Within every episode, we execute 20 actions on the deformable objects.  The data collection policy randomly selects from the action space parameters after doing collision checking and bag constraint checking, which include translations along the $x$, $y$, $z$ axes, and the rotation around $r_z$. Throughout each episode, we save the generated particle data with proposed sampling approach, as well as the robot end-effector's poses.

The experiment comprises two types of manipulation tasks. In the SOI shape preservation task, the robotic manipulators aim to keep the SOI shape consistent while moving the fabric bag from its initial shape to a predefined target shape. 
Both the initial and target configurations share the same shape but differ in their spatial transformations.
Conversely, the SOI shape servoing task involves changing the SOI shape from one configuration to another distinct shape configuration by manipulating the SOIs.
We conduct a quantitative evaluation for each component of our proposed approach, including SOI particle sampling, the SOI-based neural dynamics model, and performance results of designed manipulation tasks. 
Our primary metrics for evaluating the manipulation performance are Chamfer distance (CD) and Earth mover's distance (EMD), along with a hybrid metric that combines both. 
Furthermore, we employ Geodesic distance (GD) to assess the proximity of boundary points residing on one manifold with boundary.

\begin{table}[t]
\caption{
Averaged SOI particle sampling results across 90 frames.
}
\label{tab_sampling}
\centering
\begin{tabular}{l ccc}
 \toprule
Sampling Methods & CD(cm) $\downarrow$  & EMD(cm) $\downarrow$ & GD(cm) $\downarrow$ \\
\midrule 
Global Sampling (GPS) & $1.68 \pm 0.21$ & $1.27 \pm 0.35$ & $ 2.14 \pm 0.32$ \\
Local Sampling (LPS) & $\mathbf{1.46} \pm 0.11$ & $\mathbf{1.08} \pm 0.22$ & $\mathbf{1.97} \pm 0.23 $ \\         
\bottomrule
\end{tabular}
\end{table}

\subsection{SOI Particle Sampling}
We commence by benchmarking the proposed global particle sampling (GPS) technique against the local particle sampling (LPS) baseline. 
LPS initiates by processing the SOI-related partial point cloud of the deformable objects through its color filtering. 
Subsequently, it encapsulates the incomplete SOI point clouds with different convex hulls. 
Following this, it performs the point sampling procedure, eventually combining the sampled points into a full set of SOI particles. 
As shown in Table \ref{tab_sampling}, we compute the mean distance between the sampled particles and the ground-truth particles, the latter captured by a professional 3D scanner. Our analysis indicates that the GPS method incurs lower losses in terms of all distance metrics, thus outperforming the LPS approach. These outcomes align with the premise that incorporating additional topological information can markedly enhance the quality of sampling, particularly in scenarios where occlusions are present, and  Fig. \ref{fig_particle} presents example results for different SOIs applied to three selected deformable objects using our proposed GSP method.

 \begin{figure}[htbp]
   \vspace{-0.3cm}
    \includegraphics[width=\columnwidth]{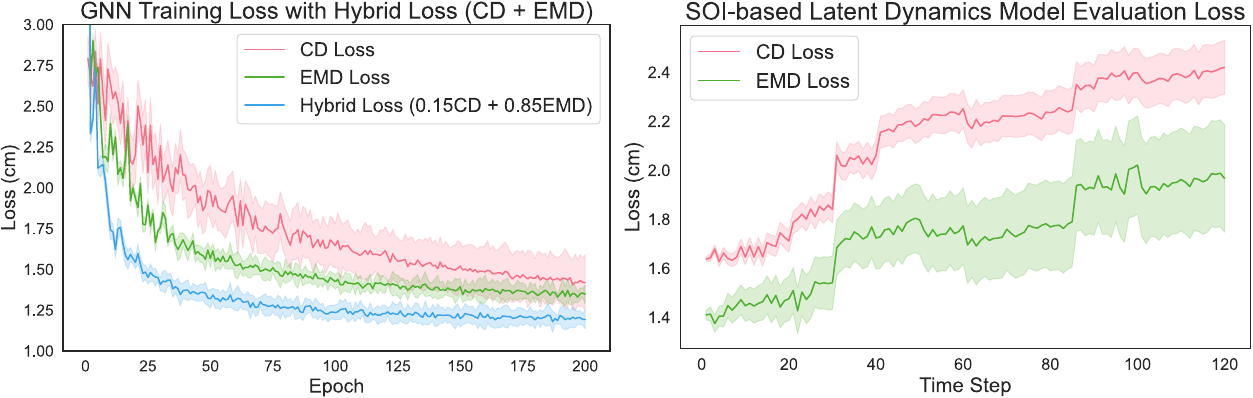}  
  \vspace{-0.6cm}
  \caption{
(Left) SOI-based latent dynamics model training loss with different loss functions in testing data.
(Right) The dynamics model's evaluation over the time horizon in testing data.
  }
  \label{fig_loss}
\end{figure}
\vspace{-0.2cm}

 \begin{figure}[htbp]
   \vspace{-0.3cm}
    \includegraphics[width=1\columnwidth]{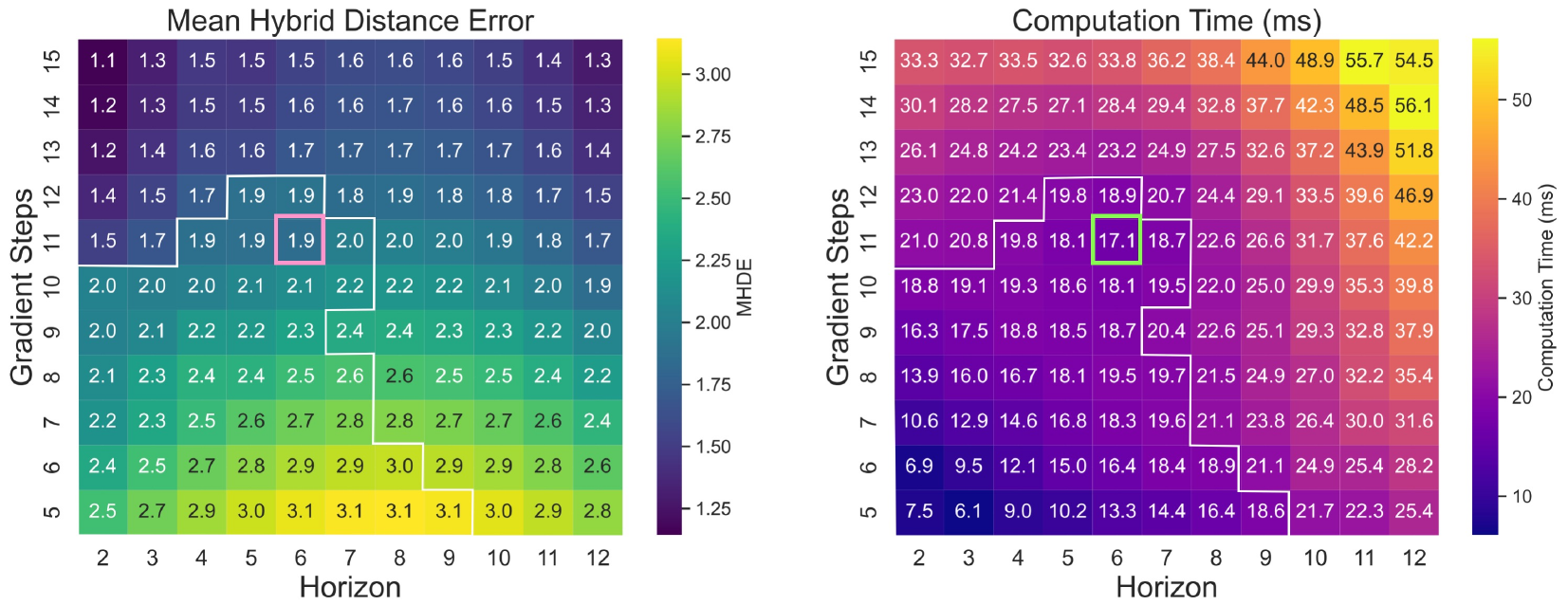}  
  \vspace{-0.8cm}
  \caption{
(Left) Mean hybrid distance error and (Right) computation time over different control horizons and gradient steps in the MPC process.
The white boundary represents the MPC computation time with respect to the horizon and gradient step being less than 20 ms, which is acceptable for the fabric manipulation task. Inside the boundary, the computation time for the specific setting of an MPC horizon of 6 and a gradient step of 11 is the most efficient.
}
  \label{fig_mpc}
\end{figure}
\vspace{-0.4cm}

\subsection{GNN-based SOI Dynamics Model}
The training of our GNN model, detailed in Section \ref{gnn_dyn}, commences with the construction of a graph, where edges are formed between vertices that are within a distance threshold of $d=0.04$. 
By adopting a 70/15/15 split, we ensure that the GNN model has enough data to learn effectively while also enabling thorough validation and testing. 
Every vertex and edge are encoded using 3-layer Multilayer Perceptrons (MLPs), featuring hidden and output layers, each with 300 neurons. 
The propagation module is constituted by a fully connected neural layer with a layer size of 300. For motion prediction, we employ an additional 3-layer MLP with the hidden layer configured to 300 neurons. 
ReLU activation functions are utilized throughout the neural networks to introduce non-linearity.
The model undergoes training for 120 epochs, employing the Adam optimizer. We have selected a batch size of 32 to balance the trade-off between generalization and computational efficiency. The learning rate is set at $5 \mathrm{e}$-4, which is a conventional choice for steady convergence. These hyperparameters were chosen to foster a robust learning process while maintaining the capacity to capture complex patterns within the data.
In practice, the final weights for the GNN-based SOI neural dynamics model are chosen based on the lowest validation loss recorded during training. This ensures that the model achieves the best possible performance on new, unseen data, which is critical for the practical application of the model in the integrated MPC for the bag manipulation task.

Subsequently, we evaluate the performance of GNN-based SOI dynamics models using different loss functions in testing data.
Table \ref{tab_loss} and Fig. \ref{fig_loss}(left) indicate that leveraging a hybrid loss function that integrates CD and EMD leads to superior performance across all metrics, as opposed to optimizing individually for either CD or EMD. Notably, the Geodesic distance is significantly improved, compared to using them individually.
The combination of $0.15$~CD + $0.85$~EMD  achieves the best performance across all metrics, giving the lowest CD, competitive EMD, and lowest GD, so we select this hybrid distance for the following experiments.
In Fig. \ref{fig_loss}(right), the dynamics model is evaluated on testing data using the CD and the EMD loss over the time horizon.
Our analysis reveals that while an unavoidable increase in loss metrics occurs as the prediction horizon of the model is extended, these metrics values remain within an acceptable range. This outcome validates the potential utility of this model in following deformable object manipulation tasks, demonstrating their capacity to maintain reasonable accuracy over a longer prediction horizon.

\begin{figure*}[htbp]
 	\centering
    \includegraphics[width=2\columnwidth]{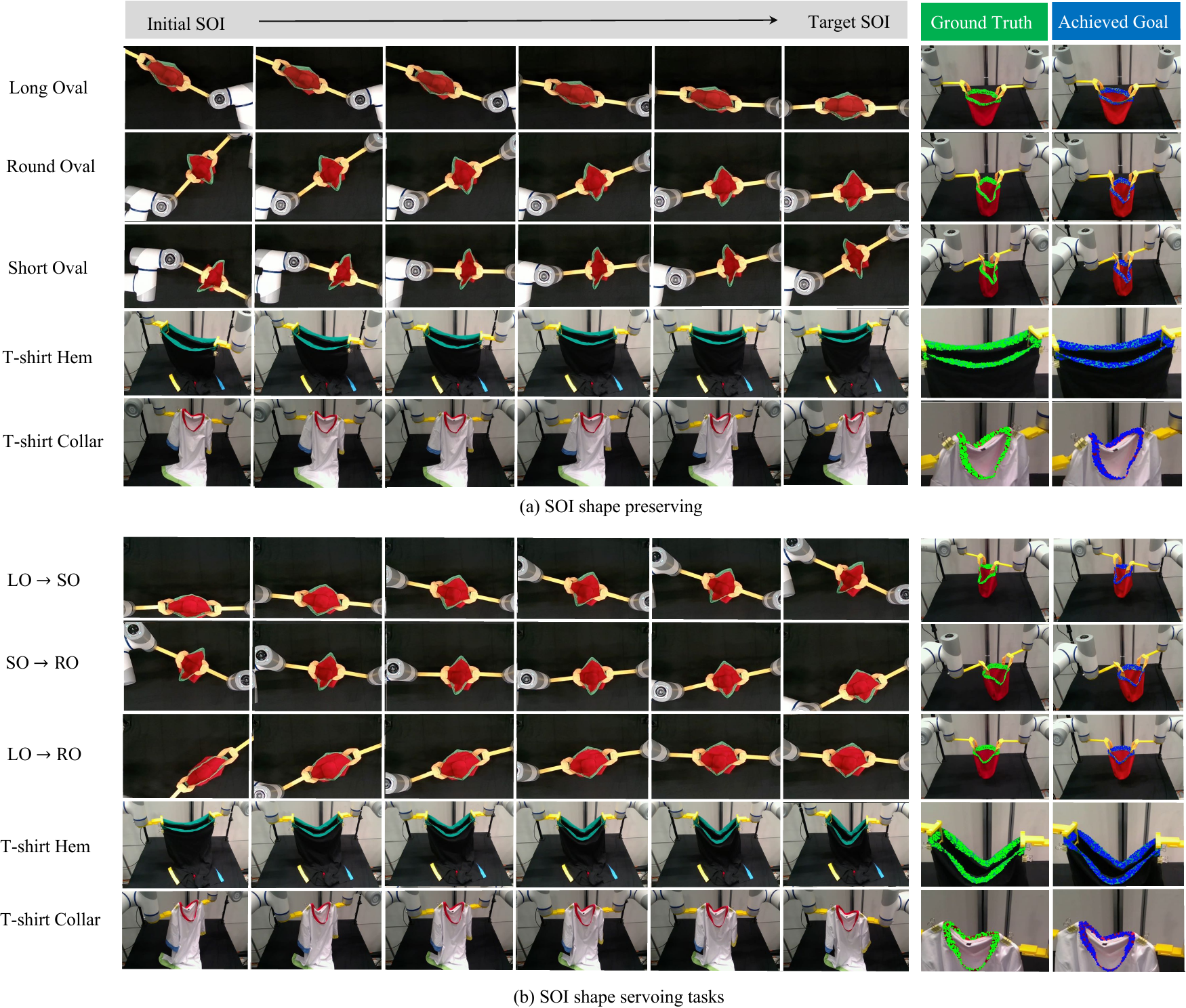}
    \vspace{-0.3cm}
  \caption{
  Qualitative results in (a) SOI shape preserving and (b) SOI shape servoing tasks with our proposed approach.  }
  \label{fig_soi_res}
   \vspace{-0.3cm}
\end{figure*}

\begin{figure*}[htbp]
	\centering
    \includegraphics[width=2\columnwidth]{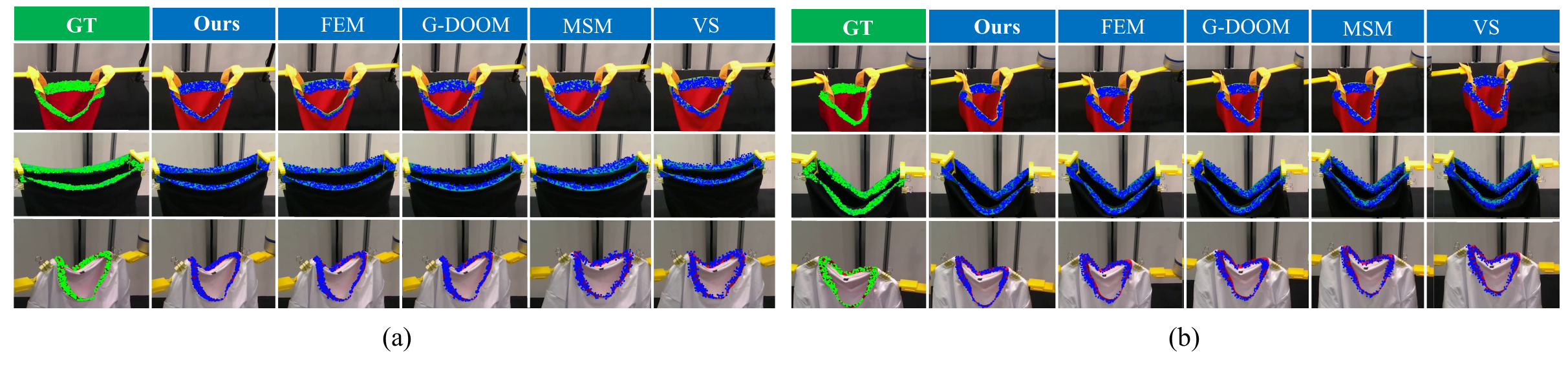}
    \vspace{-0.3cm}
  \caption{
 Achieved SOI particle configurations of (a) SOI shape preserving and (b) SOI shape servoing tasks by different approaches.
  }
  \label{fig_soi_res2}
   \vspace{-0.3cm}
\end{figure*}

To evaluate the performance of the proposed SOI-based neural dynamics model for real deformable object manipulation tasks, we integrate it with MPC to conduct SOI-preserving experiments on a fabric bag with various oval shapes.
To balance real-time performance and control accuracy for our manipulation task, we set the MPC horizon to 6 and the gradient step to 11, based on extensive empirical testing (see Fig. \ref{fig_mpc}). This combination provides a reasonable compromise between computational demand and the quality of optimization for the MPC. 
We compare the performance of our proposed approach with two commonly-used and well-established dynamics modeling techniques: the Mass-Spring Model (MSM) \cite{makiyeh2022indirect} and the Finite Element Model (FEM) \cite{zhang2017visual}.
These two molding techniques are also integrated into the MPC framework for conducting the experiments.
The comparative analysis over 60 trails includes both quantitative and qualitative outcomes, as shown in Table \ref{tab_soi_perserve},  Fig. \ref{fig_soi_res}(a) and Fig. \ref{fig_soi_res2}(a). 
Our findings reveal a clear trend across all examined methods; there is an incremental rise in error corresponding to the transformation of the bag's shape from a Long Oval (LO) to a Short Oval (SO), suggesting that as the bag opening rim becomes shorter and wider, the complexity of preserving its structure increases. Notably, our GNN-based dynamics model consistently outperforms the MSM and FEM across all object shapes, achieving the lowest hybrid loss in tasks dedicated to shape preservation. This outcome indicates the superior expressive capability of our GNN-based model in capturing the intricate deformations of the fabric bags, and maintaining the SOI shape unchanged during the fabric bag moving, surpassing the traditional MSM and FEM approaches, particularly when dealing with objects that exhibit more complex dynamic behaviors. The advantage of our model is most pronounced when interacting with simpler geometric shapes.


\begin{table}[t]
\caption{
The performance of the SOI particle dynamics model with different loss functions in testing data
}
\label{tab_loss}
\centering
\begin{tabular}{l cccc}
 \toprule
Loss Functions & CD(cm) $\downarrow$ & EMD(cm) $\downarrow$ & GD(cm) $\downarrow$ \\ 
\midrule 
CD                      & $1.52 \pm 0.19$  & $1.94 \pm 0.18$ & $2.53 \pm 0.25$ \\
EMD                     & $1.85 \pm 0.17$  & $\mathbf{1.37} \pm 0.15$ & $2.76 \pm 0.13$ \\
$0.2$~CD + $0.8$~EMD    & $1.59 \pm 0.16$  & $1.40 \pm 0.17$ & $2.13 \pm 0.18$ \\
$0.1$~CD + $0.9$~EMD    & $1.55 \pm 0.17$  & $1.43 \pm 0.16$ & $2.28 \pm 0.19$ \\
$0.15$~CD + $0.85$~EMD  & $\mathbf{1.50} \pm 0.16$  & $1.38 \pm 0.17$ & $ \mathbf{2.02} \pm 0.16$ \\    
\bottomrule
\end{tabular}
\end{table}

\begin{table}[t]
\caption{
Mean Hybrid Distance Error for SOI Shape Preserving by Different Dynamics Modeling Methods over 60 Trials}
\label{tab_soi_perserve}
\centering
\begin{tabular}{l cccc}
 \toprule
Method &  Long Oval (LO) & Round Oval (RO) & Short Oval (SO) \\            
 \midrule       
MSM	&  \(2.63 \pm 0.31\)  & \(2.82 \pm 0.38\) & \(2.94 \pm 0.41\)  \\
FEM	 & \(1.96 \pm 0.27\) & \(2.13 \pm 0.36\) & \(2.42 \pm 0.38\)  \\
Ours & \(\textbf{1.69} \pm 0.18\) & \(\textbf{1.84} \pm 0.22 \) & \(\textbf{2.08} \pm 0.21\)  \\
\bottomrule
\end{tabular}
\vspace{-0.4cm}
\end{table}

\begin{table*}
\caption{
Mean Hybrid Distance Error and Success Rate for SOI Shape Preserving on Different Deformable Objects by Different Methods }
\label{tab_soi_preserve}
\centering
\begin{tabular}{l cc cc cc c}
 \toprule
\multirow{2}{*}{Method} & \multicolumn{2}{c}{Bag Opening Rim (180 trls)} & \multicolumn{2}{c}{Soft T-shirt Collar (60 trls)} & \multicolumn{2}{c}{Stiff T-shirt Bottom Hem (60 trls)} & \multirow{2}{*}{\textit{Total}} \\
\cmidrule(lr){2-3}
\cmidrule(lr){4-5}
\cmidrule(lr){6-7}
 ~ & Mean HD Error $\downarrow$ &  Success Rate $\uparrow$ &  Mean HD Error $\downarrow$ &  Success Rate $\uparrow$ &  Mean HD Error $\downarrow$ &  Success Rate $\uparrow$ \\            
 \midrule       
VS	\cite{lagneau2020active} & \(3.20 \pm 0.20\)  & \(71.11\%\) & \(3.41 \pm 0.21\)  & \(76.67\%\) & \(3.28 \pm 0.29\)  & \(68.33\%\) & \(71.67\%\) \\
MSM \cite{makiyeh2022indirect} & \(2.80 \pm 0.37\)  & \(73.89\%\) & \(2.48 \pm 0.40\)  & \(71.67\%\) & \(3.06 \pm 0.48\)  & \(65.00\%\) & \(71.67\%\) \\
FEM \cite{zhang2017visual} & \(2.17 \pm 0.34\)  & \(83.33\%\) & \(2.23 \pm 0.33\)  & \(80.00\%\) & \(2.64 \pm 0.38\)  & \(86.67\%\) & \(83.33\%\) \\
G-DOOM \cite{ma2022learning} & \(3.07 \pm 0.32\)  & \(79.44\%\) & \(2.64 \pm 0.34 \)  & \(75.00\%\) & \(3.31 \pm 0.43\)  & \(71.67\%\) & \(77.00\%\) \\
\textbf{Ours} & \(1.86 \pm 0.23\)  & \(\mathbf{98.33\%}\) & \(1.65 \pm 0.27\)  & \(\mathbf{95.00\%}\) & \(2.43 \pm 0.26\)  & \(\mathbf{86.67\%}\) & \(\mathbf{95.33\%}\) \\
\bottomrule
\end{tabular}
\end{table*}

\begin{table*}
\caption{
Mean Hybrid Distance Error and Success Rate for SOI Shape Servoing on Different Deformable Objects by Different Methods }
\label{tab_soi_servo}
\centering
\begin{tabular}{l cc cc cc c}
 \toprule
\multirow{2}{*}{Method} & \multicolumn{2}{c}{Bag Opening Rim (180 trls)} & \multicolumn{2}{c}{Soft T-shirt Collar (60 trls)} & \multicolumn{2}{c}{Stiff T-shirt Bottom Hem (60 trls)} & \multirow{2}{*}{\textit{Total}} \\
\cmidrule(lr){2-3}
\cmidrule(lr){4-5}
\cmidrule(lr){6-7}
 ~ & Mean HD Error $\downarrow$ &  Success Rate $\uparrow$ &  Mean HD Error $\downarrow$ &  Success Rate $\uparrow$ &  Mean HD Error $\downarrow$ &  Success Rate $\uparrow$ \\            
 \midrule       
VS	\cite{lagneau2020active} & \(3.76 \pm 0.21\)  & \(68.89\%\) & \(3.75 \pm 0.19\)  & \(71.67\%\) & \(3.84 \pm 0.27\)  & \(70.00\%\) & \(69.67\%\) \\
MSM \cite{makiyeh2022indirect} & \(3.29 \pm 0.45\)  & \(70.00\%\) & \(2.94 \pm 0.42\)  & \(66.67\%\) & \(3.57 \pm 0.47\)  & \(63.33\%\) & \(68.00\%\) \\
FEM \cite{zhang2017visual} & \(2.70 \pm 0.35\)  & \(78.89\%\) & \(2.62 \pm 0.37\)  & \(76.67\%\) & \(2.88 \pm 0.39\)  & \(80.00\%\) & \(78.67\%\) \\
G-DOOM \cite{ma2022learning} & \(3.48 \pm 0.33\)  & \(75.56\%\) & \(3.11 \pm 0.37 \)  & \(68.33\%\) & \(3.87 \pm 0.41\)  & \(66.67\%\) & \(72.33\%\) \\
\textbf{Ours} & \(2.27 \pm 0.24\)  & \(\mathbf{96.67\%}\) & \(1.87 \pm 0.26\)  & \(\mathbf{91.67\%}\) & \(2.68 \pm 0.29\)  & \(\mathbf{90.00\%}\) & \(\mathbf{94.33\%}\) \\
\bottomrule
\end{tabular}
\end{table*}

\subsection{Manipulation Results}
To extensively validate our proposed deformable object manipulation framework using the SOI-based neural dynamics model-integrated MPC approach, we specifically selected two additional T-shirts with different stiffness characteristics to demonstrate the effectiveness of our method. One SOI is the collar of a T-shirt made of soft modal fabric, known for its lightweight and stretchy properties. The other SOI is the bottom hem of a T-shirt made of a polyester-cotton blend, which is notably stiffer in comparison.
Additionally, we compare our manipulation framework with a popular shape servoing technique tailored for deformable objects--visual servoing (VS) as proposed by Lagneau \textit{et al.} \cite{lagneau2020active}, and an elaborately designed GNN-based latent graph dynamics model--G-DOOM as introduced by Ma \textit{et al.} for a comprehensive comparison analysis. 
The analysis integrates both quantitative and qualitative assessments, as depicted in Fig. \ref{fig_soi_res} and Fig. \ref{fig_soi_res2} and detailed in Table \ref{tab_soi_preserve} and Table \ref{tab_soi_servo}. 
In shape servoing experiments, we consider five specific shape servoing tasks: LO$\rightarrow$SO, SO$\rightarrow$RO, and LO$\rightarrow$RO for fabric bag, LO$\rightarrow$RO for the bottom hem of a T-shirt, and servoing the collar of a T-shirt. The evaluation metrics include the Mean Hybrid Distance Error (recorded solely for successful trials) and the Success Rate of SOI servoing. Furthermore, we report the success rate for achieving complete servoing with a threshold error of less than $5$ cm, based on 60 trials for each method and SOI servoing case. The experimental outcomes reveal that our proposed method consistently secured the least shape error, ranging from $1.87$ to $2.68$ cm across all tested cases, and achieved an outstanding overall success rate of $94.33\%$.  When contrasted with traditional model-based techniques such as the Mass-Spring Model and the Finite Element Model, our SOI-based latent dynamics model exhibited a markedly enhanced capability in manipulating the SOI for the fabric bag and the T-shirt. The G-DOOM approach underperforms due to its reliance on sparse key point extraction via Transporter Networks, providing insufficient data for Model Predictive Control (MPC) to accurately guide robotic motion. However, our approach is based on a dense particle set that offers more detailed information, enabling the MPC to optimize movements more effectively for manipulating deformable objects. It is observed that shorter SOI shapes lead to lower errors and longer SOI shapes result in larger errors. Nevertheless, our proposed method demonstrated a robust performance across all tasks.

\section{Conclusion}
In this work, we introduced a novel bimanual manipulation framework for deformable bags, centered around a SOI-based latent dynamics model. Our approach effectively integrates multi-view perception, graph neural networks, and model predictive control, facilitating precise and efficient robotic manipulation of flexible materials. Through real-world experiments, the framework demonstrated promising results in intelligent physical interaction with deformable objects.
One limitation of the current system is its reliance on differently colored SOIs to distinguish target manipulation areas, which may not be feasible in all operational settings. Future efforts will focus on employing a learning-based SOI detection module and multimodal sensing to enhance the system's ability to identify and manipulate SOIs without such visual aids, thereby broadening the applicability of our method to a wider array of real-world scenarios.

\bibliographystyle{IEEEtran}
\bibliography{refs}

\begin{thebibliography}{10}
\providecommand{\url}[1]{#1}
\csname url@samestyle\endcsname
\providecommand{\newblock}{\relax}
\providecommand{\bibinfo}[2]{#2}
\providecommand{\BIBentrySTDinterwordspacing}{\spaceskip=0pt\relax}
\providecommand{\BIBentryALTinterwordstretchfactor}{4}
\providecommand{\BIBentryALTinterwordspacing}{\spaceskip=\fontdimen2\font plus
\BIBentryALTinterwordstretchfactor\fontdimen3\font minus \fontdimen4\font\relax}
\providecommand{\BIBforeignlanguage}[2]{{%
\expandafter\ifx\csname l@#1\endcsname\relax
\typeout{** WARNING: IEEEtran.bst: No hyphenation pattern has been}%
\typeout{** loaded for the language `#1'. Using the pattern for}%
\typeout{** the default language instead.}%
\else
\language=\csname l@#1\endcsname
\fi
#2}}
\providecommand{\BIBdecl}{\relax}
\BIBdecl

\bibitem{yin2021modeling}
H.~Yin, A.~Varava, and D.~Kragic, ``Modeling, learning, perception, and control methods for deformable object manipulation,'' \emph{{Sci. Robot.}}, vol.~6, no.~54, 2021.

\bibitem{zhu2022challenges}
J.~Zhu, A.~Cherubini, C.~Dune, D.~Navarro-Alarcon, F.~Alambeigi, D.~Berenson, F.~Ficuciello, K.~Harada, J.~Kober, X.~Li \emph{et~al.}, ``Challenges and outlook in robotic manipulation of deformable objects,'' \emph{{IEEE Robot. Autom. Mag.}}, vol.~29, no.~3, pp. 67--77, 2022.

\bibitem{hu20193}
Z.~Hu, T.~Han, P.~Sun, J.~Pan, and D.~Manocha, ``3-d deformable object manipulation using deep neural networks,'' \emph{{IEEE Robot. Autom. Lett.}}, vol.~4, no.~4, pp. 4255--4261, 2019.

\bibitem{garcia2022household}
I.~Garcia-Camacho, J.~Borr{\`a}s, B.~Calli, A.~Norton, and G.~Aleny{\`a}, ``Household cloth object set: Fostering benchmarking in deformable object manipulation,'' \emph{{IEEE Robot. Autom. Lett.}}, vol.~7, no.~3, pp. 5866--5873, 2022.

\bibitem{huo2022keypoint}
S.~Huo, A.~Duan, C.~Li, P.~Zhou, W.~Ma, H.~Wang, and D.~Navarro-Alarcon, ``Keypoint-based planar bimanual shaping of deformable linear objects under environmental constraints with hierarchical action framework,'' \emph{{IEEE Robot. Autom. Lett.}}, vol.~7, no.~2, pp. 5222--5229, 2022.

\bibitem{lin2022diffskill}
X.~Lin, Z.~Huang, Y.~Li, J.~B. Tenenbaum, D.~Held, and C.~Gan, ``Diffskill: Skill abstraction from differentiable physics for deformable object manipulations with tools,'' in \emph{{Proc. Int. Conf. Learn. Represent.}}, 2022.

\bibitem{makiyeh2022indirect}
F.~Makiyeh, M.~Marchal, F.~Chaumette, and A.~Krupa, ``Indirect positioning of a 3d point on a soft object using rgb-d visual servoing and a mass-spring model,'' in \emph{2022 17th International Conference on Control, Automation, Robotics and Vision (ICARCV)}.\hskip 1em plus 0.5em minus 0.4em\relax IEEE, 2022, pp. 235--242.

\bibitem{zhang2017visual}
Z.~Zhang, T.~M. Bieze, J.~Dequidt, A.~Kruszewski, and C.~Duriez, ``Visual servoing control of soft robots based on finite element model,'' in \emph{{IEEE/RSJ Int. Conf. on Robots and Intelligent Systems}}.\hskip 1em plus 0.5em minus 0.4em\relax IEEE, 2017, pp. 2895--2901.

\bibitem{ficuciello2018fem}
F.~Ficuciello, A.~Migliozzi, E.~Coevoet, A.~Petit, and C.~Duriez, ``Fem-based deformation control for dexterous manipulation of 3d soft objects,'' in \emph{{IEEE/RSJ Int. Conf. on Robots and Intelligent Systems}}.\hskip 1em plus 0.5em minus 0.4em\relax IEEE, 2018, pp. 4007--4013.

\bibitem{xu2022dextairity}
Z.~Xu, C.~Chi, B.~Burchfiel, E.~Cousineau, S.~Feng, and S.~Song, ``Dextairity: Deformable manipulation can be a breeze,'' in \emph{Proceedings of Robotics: Science and Systems (RSS)}, 2022.

\bibitem{zhou2024reactive}
P.~Zhou, P.~Zheng, J.~Qi, C.~Li, H.-Y. Lee, A.~Duan, L.~Lu, Z.~Li, L.~Hu, and D.~Navarro-Alarcon, ``Reactive human--robot collaborative manipulation of deformable linear objects using a new topological latent control model,'' \emph{{Robot. Comput.-Integr. Manuf.}}, vol.~88, p. 102727, 2024.

\bibitem{navarro2016automatic}
D.~Navarro-Alarcon, H.~M. Yip, Z.~Wang, Y.-H. Liu, F.~Zhong, T.~Zhang, and P.~Li, ``Automatic 3-d manipulation of soft objects by robotic arms with an adaptive deformation model,'' \emph{{IEEE Trans. Robot.}}, vol.~32, no.~2, pp. 429--441, 2016.

\bibitem{qi2021contour}
J.~Qi, G.~Ma, J.~Zhu, P.~Zhou, Y.~Lyu, H.~Zhang, and D.~Navarro-Alarcon, ``Contour moments based manipulation of composite rigid-deformable objects with finite time model estimation and shape/position control,'' \emph{{IEEE/ASME Trans. Mechatron.}}, vol.~27, no.~5, pp. 2985--2996, 2021.

\bibitem{lin2022planning}
X.~Lin, C.~Qi, Y.~Zhang, Z.~Huang, K.~Fragkiadaki, Y.~Li, C.~Gan, and D.~Held, ``Planning with spatial-temporal abstraction from point clouds for deformable object manipulation,'' in \emph{{Conf. Rob. Learn.}}, 2022.

\bibitem{nair2017combining}
A.~Nair, D.~Chen, P.~Agrawal, P.~Isola, P.~Abbeel, J.~Malik, and S.~Levine, ``Combining self-supervised learning and imitation for vision-based rope manipulation,'' in \emph{{IEEE Int. Conf. on Robotics and Automation}}.\hskip 1em plus 0.5em minus 0.4em\relax IEEE, 2017, pp. 2146--2153.

\bibitem{yan2020self}
M.~Yan, Y.~Zhu, N.~Jin, and J.~Bohg, ``Self-supervised learning of state estimation for manipulating deformable linear objects,'' \emph{{IEEE Robot. Autom. Lett.}}, vol.~5, no.~2, pp. 2372--2379, 2020.

\bibitem{gasteiger2021gemnet}
J.~Gasteiger, F.~Becker, and S.~G{\"u}nnemann, ``Gemnet: Universal directional graph neural networks for molecules,'' \emph{{Proc. Adv. Neural Inf. Process. Syst.}}, vol.~34, pp. 6790--6802, 2021.

\bibitem{zhou2024imitating}
P.~Zhou, J.~Qi, A.~Duan, S.~Huo, Z.~Wu, and D.~Navarro-Alarcon, ``Imitating tool-based garment folding from a single visual observation using hand-object graph dynamics,'' \emph{{IEEE Trans. Ind. Informat.}}, 2024.

\bibitem{tolstaya2020learning}
E.~Tolstaya, F.~Gama, J.~Paulos, G.~Pappas, V.~Kumar, and A.~Ribeiro, ``Learning decentralized controllers for robot swarms with graph neural networks,'' in \emph{{Conf. Rob. Learn.}}\hskip 1em plus 0.5em minus 0.4em\relax PMLR, 2020, pp. 671--682.

\bibitem{bertiche2022neural}
H.~Bertiche, M.~Madadi, and S.~Escalera, ``Neural cloth simulation,'' \emph{{ACM Trans. Graph.}}, vol.~41, no.~6, pp. 1--14, 2022.

\bibitem{wang2022offline}
C.~Wang, Y.~Zhang, X.~Zhang, Z.~Wu, X.~Zhu, S.~Jin, T.~Tang, and M.~Tomizuka, ``Offline-online learning of deformation model for cable manipulation with graph neural networks,'' \emph{{IEEE Robot. Autom. Lett.}}, vol.~7, no.~2, pp. 5544--5551, 2022.

\bibitem{deng2022deep}
Y.~Deng, C.~Xia, X.~Wang, and L.~Chen, ``Deep reinforcement learning based on local gnn for goal-conditioned deformable object rearranging,'' in \emph{2022 IEEE/RSJ International Conference on Intelligent Robots and Systems (IROS)}.\hskip 1em plus 0.5em minus 0.4em\relax IEEE, 2022, pp. 1131--1138.

\bibitem{shi2023robocraft}
H.~Shi, H.~Xu, Z.~Huang, Y.~Li, and J.~Wu, ``Robocraft: Learning to see, simulate, and shape elasto-plastic objects in 3d with graph networks,'' \emph{{Int. J. Robot. Res.}}, p. 02783649231219020, 2023.

\bibitem{ma2022learning}
X.~Ma, D.~Hsu, and W.~S. Lee, ``Learning latent graph dynamics for visual manipulation of deformable objects,'' in \emph{2022 International Conference on Robotics and Automation (ICRA)}.\hskip 1em plus 0.5em minus 0.4em\relax IEEE, 2022, pp. 8266--8273.

\bibitem{gu2024shakingbot}
N.~Gu, Z.~Zhang, R.~He, and L.~Yu, ``Shakingbot: dynamic manipulation for bagging,'' \emph{Robotica}, vol.~42, no.~3, pp. 775--791, 2024.

\bibitem{chen2023autobag}
L.~Y. Chen, B.~Shi, D.~Seita, R.~Cheng, T.~Kollar, D.~Held, and K.~Goldberg, ``Autobag: Learning to open plastic bags and insert objects,'' in \emph{{IEEE Int. Conf. on Robotics and Automation}}.\hskip 1em plus 0.5em minus 0.4em\relax IEEE, 2023, pp. 3918--3925.

\bibitem{weng2024interactive}
Z.~Weng, P.~Zhou, H.~Yin, A.~Kravberg, A.~Varava, D.~Navarro-Alarcon, and D.~Kragic, ``Interactive perception for deformable object manipulation,'' \emph{arXiv preprint arXiv:2403.05177}, 2024.

\bibitem{liu1989limited}
D.~C. Liu and J.~Nocedal, ``On the limited memory bfgs method for large scale optimization,'' \emph{Mathematical programming}, vol.~45, no.~1, pp. 503--528, 1989.

\bibitem{mokhtari2015global}
A.~Mokhtari and A.~Ribeiro, ``Global convergence of online limited memory bfgs,'' \emph{{J. Mach. Learn. Res.}}, vol.~16, no.~1, pp. 3151--3181, 2015.

\bibitem{lagneau2020active}
R.~Lagneau, A.~Krupa, and M.~Marchal, ``Active deformation through visual servoing of soft objects,'' in \emph{{IEEE Int. Conf. on Robotics and Automation}}.\hskip 1em plus 0.5em minus 0.4em\relax IEEE, 2020, pp. 8978--8984.

\end{thebibliography}


\begin{IEEEbiography}
[{\includegraphics[width=1in,height=1.25in,clip,keepaspectratio]{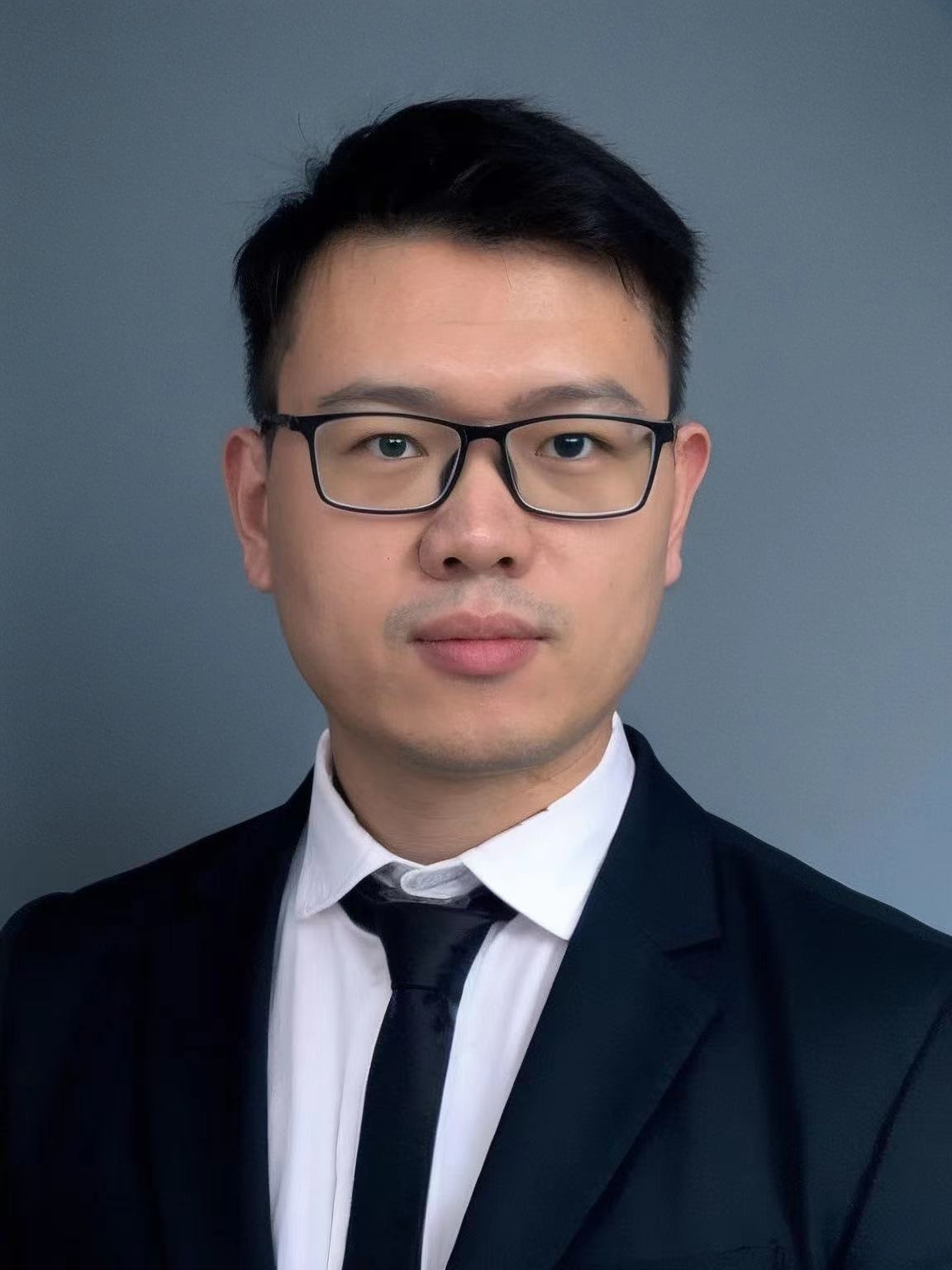}}]
{Peng Zhou} 
received the M.Sc. degree in software engineering from Tongji University, Shanghai, China, in 2017, and Ph.D. degree in robotics from The Hong Kong Polytechnic University, Hong Kong SAR, in 2022.
In 2021, he was a visiting Ph.D. student at Robotics, Perception and Learning Lab, KTH Royal Institute of Technology, Stockholm, Sweden. 
He is currently a Postdoctoral Research Fellow at The University of Hong Kong. His research interests include deformable object manipulation, robot perception and learning, and task and motion planning.
\end{IEEEbiography}
\vspace{-1.5cm}

\begin{IEEEbiography} [{\includegraphics[width=1in,height=1.25in,clip,keepaspectratio]{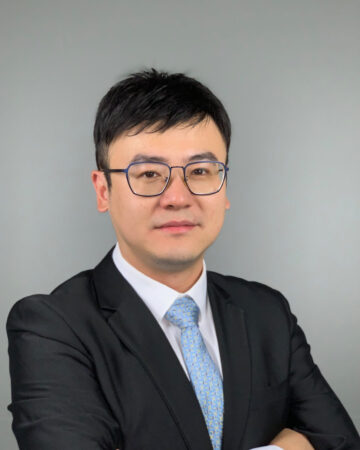}}]
{Pai Zheng}  (Senior Member, IEEE) received the dual bachelor’s degrees in mechanical engineering (major) and computer science and engineering (minor) from the Huazhong University of Science and Technology, Wuhan, China, in 2010, the master’s degree in mechanical engineering from Beihang University, Beijing, China, in 2013, and the Ph.D. degree in mechanical engineering from The University of Auckland, Auckland, New Zealand, in 2017. He is currently an Assistant Professor with the Department of Industrial and Systems Engineering, The Hong Kong Polytechnic University. His research interests include smart product-service systems, human–robot collaboration, and smart manufacturing systems. He is a member of HKIE, CMES, and ASME.
He serves as an Associate Editor for Journal of Intelligent Manufacturing and Journal of Cleaner Production, an Editorial Board Member for the Journal of Manufacturing Systems and Advanced Engineering Informatics, and a guest editor/reviewer for several high impact international journals in the manufacturing and industrial engineering field.
\end{IEEEbiography}
\vspace{-1.5cm}

\begin{IEEEbiography}
[{\includegraphics[width=1in,height=1.25in,clip,keepaspectratio]{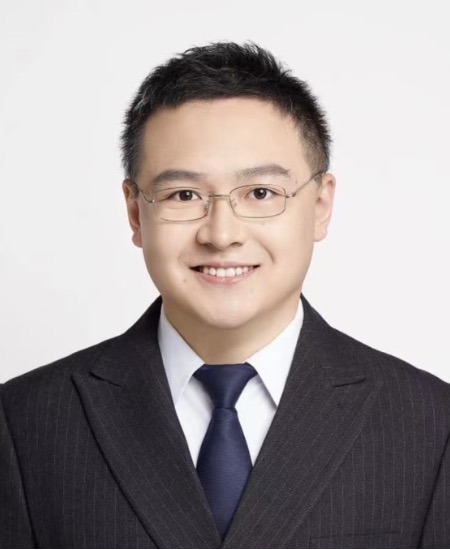}}]{Jiaming Qi} 
received the Ph.D degree in control science and engineering, and the M.S. degree in Integrated Circuit Engineering from the Harbin Institute of Technology, Harbin, China, in 2023 and 2018, respectively. 
He performs research in the deformable object manipulation, visual servoing, and human-robot collaboration.
He is currently a post-doctoral fellow in the Centre for Transformative Garment Production, The University of Hong Kong, Hong Kong.
\end{IEEEbiography}
\vspace{-1.5cm}

\begin{IEEEbiography}
[{\includegraphics[width=1in,height=1.25in,clip,keepaspectratio]{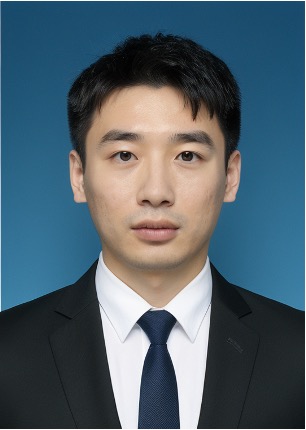}}]{Chengxi Li}  (Graduate Student Member, IEEE) received the B.E. degree in Information Technology from Vaasa University of Applied Sciences, Finland in 2018, and the M.S. degree in Computer Science from Uppsala University, Sweden in 2020, respectively. He is currently pursuing the Ph.D. degree with the Department of Industrial and Systems Engineering, The Hong Kong Polytechnic University, China. His research interests include robot learning, mixed reality, and human–robot collaboration.
\end{IEEEbiography}
\vspace{-1cm}

\begin{IEEEbiography}[{\includegraphics[width=1in,height=1.25in,clip,keepaspectratio]{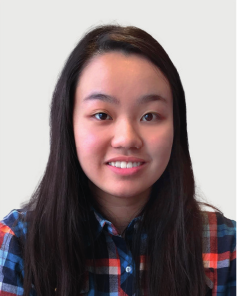}}] 
    {Hoi-Yin Lee} received her B.Eng. degree in Mechanical Engineering from The Hong Kong Polytechnic University of Hong Kong (PolyU), Hong Kong, in 2021. She was a visiting scholar at Bristol Robotics Laboratory, United Kingdom, in 2024.
    She is currently pursuing her Ph.D. degree in Mechanical Engineering at PolyU. Her research interests include tool manipulation, multi-robot systems, perceptual robots, image processing, and automation.
\end{IEEEbiography}
\vspace{-1cm}

\begin{IEEEbiography}[{\includegraphics[width=1in,height=1.25in,clip,keepaspectratio]{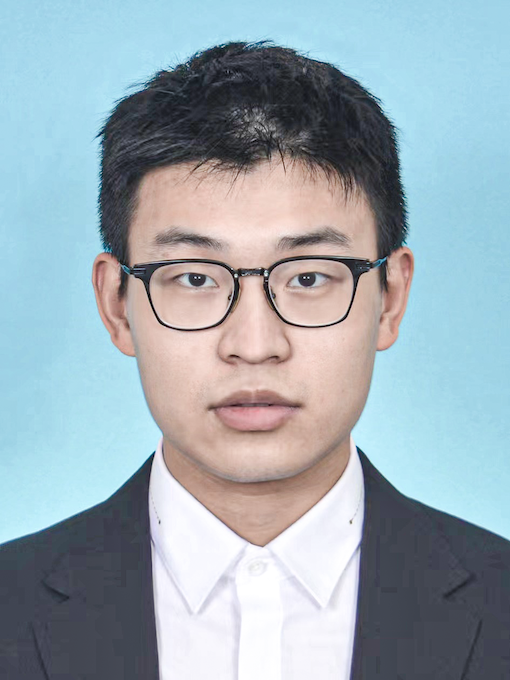}}] 
{Yipeng Pan} received the B.Eng. degree in Electronic Engineering from the China University of Mining and Technology, Xuzhou, China, in 2019. He is currently pursuing the Ph.D. degree with the Department of Computer Science, The University of Hong Kong, China. His research interests include novel sensors, and sensor fusion.
\end{IEEEbiography}
\vspace{-1cm}

\begin{IEEEbiography}
[{\includegraphics[width=1in,height=1.25in,clip,keepaspectratio]{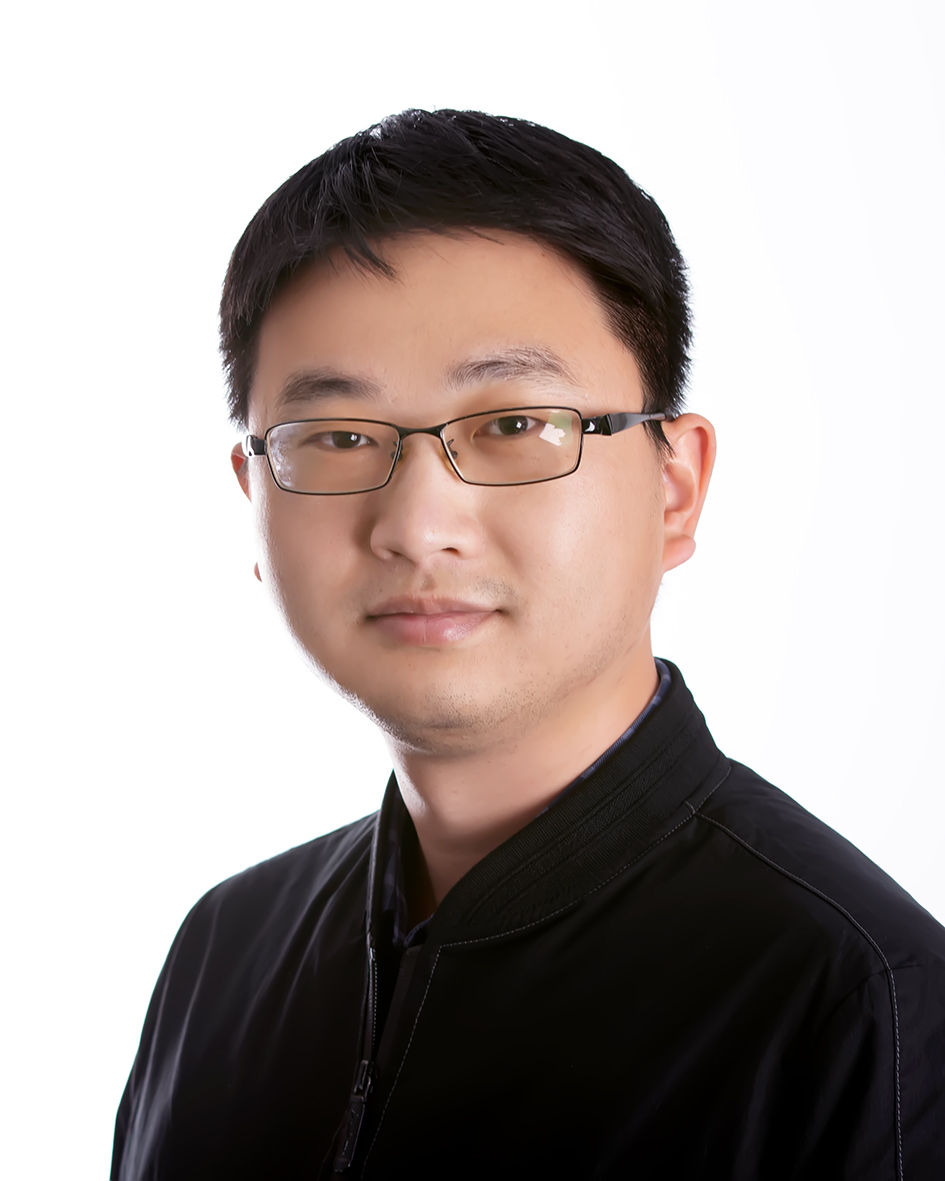}}] {Chenguang Yang} (Fellow, IEEE)  received the B.Eng. degree in measurement and control from Northwestern Polytechnical University, Xian, China, in 2005, and the Ph.D. degree in control engineering from the National University of Singapore, Singapore, in 2010. He performed postdoctoral studies in human robotics at the Imperial College London, London, U.K from 2009 to 2010.  He is Chair in Robotics with Department of Computer Science, University of Liverpool, UK.  He was awarded UK EPSRC UKRI Innovation Fellowship and individual EU Marie Curie International Incoming Fellowship. As the lead author, he won the IEEE Transactions on Robotics Best Paper Award (2012) and IEEE Transactions on Neural Networks and Learning Systems Outstanding Paper Award (2022). He is the Corresponding Co-Chair of IEEE Technical Committee on Collaborative Automation for Flexible Manufacturing. His research interest lies in human robot interaction and intelligent system design.
\end{IEEEbiography}
\vspace{-1cm}

\begin{IEEEbiography}
[{\includegraphics[width=1in,height=1.25in,clip,keepaspectratio]{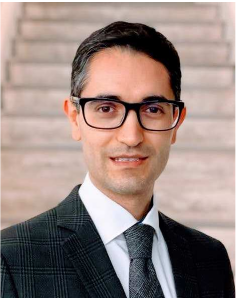}}]  {David Navarro-Alarcon} (Senior Member, IEEE) received his Ph.D. degree in mechanical and automation engineering from The Chinese University of Hong Kong in 2014.
    He is currently an Associate Professor at the Department of Mechanical Engineering at The Hong Kong Polytechnic University (PolyU). His current research interests include perceptual robotics and control theory. He currently serves as an Associate Editor of the \textsc{IEEE Transactions on Robotics}.
\end{IEEEbiography}
\vspace{-1cm}

\begin{IEEEbiography}
[{\includegraphics[width=1in,height=1.25in,clip,keepaspectratio]{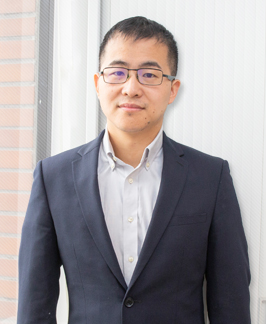}}] {Jia Pan} (Senior Member, IEEE) received the Ph.D. degree in computer science from the University of North Carolina at Chapel Hill, Chapel Hill, NC, USA, in 2013.

He is currently an Associate Professor with the Department of Computer Science, University of Hong Kong, Hong Kong. He is also a member of the Centre for Garment Production Limited, Hong Kong. His research interests include robotics and artificial intelligence as applied to autonomous systems, particularly for navigation and manipulation in challenging tasks such as effective movement in dense human crowds and manipulating deformable objects for garment automation.
\end{IEEEbiography}

\vfill

\end{document}